\documentclass[manuscript,screen, review=false]{acmart}

\AtBeginDocument{%
  \providecommand\BibTeX{{%
    \normalfont B\kern-0.5em{\scshape i\kern-0.25em b}\kern-0.8em\TeX}}}
\newcommand\blfootnote[1]{%
  \begingroup
  \renewcommand\thefootnote{}\footnote{#1}%
  \addtocounter{footnote}{-1}%
  \endgroup
}

\setcopyright{acmcopyright}
\copyrightyear{2020}
\acmYear{2020}
  \acmDOI{10.1145/1122445.1122456}

 \usepackage{multirow}  

\begin{document}

\title{A Comprehensive Review on Recent Methods and Challenges of Video Description }
\author{Alok Singh}
\email{alok_rs@cse.nits.ac.in}
\orcid{0000-0002-2683-0542}

\author{Thoudam Doren Singh}
\email{doren@cse.nits.ac.in}
\orcid{0000-0001-9906-9136}

\author{Sivaji Bandyopadhyay}
\email{sivaji.cse.ju@gmail.com}
\affiliation{%
  \institution{Centre for Natural Language Processing (CNLP) and Department of Computer Science and Engineering, National Institute of Technology Silchar, India}
  \city{Silchar}
  \state{Assam}
  \postcode{788010}
}
\renewcommand{\shortauthors}{Singh, et. al.}

\begin{abstract}
Video description involves the generation of the natural language description of actions, events, and objects in the video. There are various applications of video description by filling the gap between languages and vision for visually impaired people, generating automatic title suggestion based on content, browsing of the video based on the content and video-guided machine translation \cite{wang2019vatex} etc. In the past decade, several works had been done in this field in terms of approaches/methods for video description, evaluation metrics, and datasets. For analyzing the progress in the video description task, a comprehensive survey is needed that covers all the phases of video description approaches with a special focus on recent deep learning approaches. In this work, we report a comprehensive survey on the phases of video description approaches, the dataset for video description, evaluation metrics, open competitions for motivating the research on the video description, open challenges in this field, and future research directions. In this survey, we cover the state-of-the-art approaches proposed for each and every dataset with their pros and cons. For the growth of this research domain, the availability of numerous benchmark dataset is a basic need. Further, we categorize all the dataset into two classes: open domain dataset and domain-specific dataset. A brief discussion of the pros and cons of automatic evaluation metrics and human evaluation is also done in this survey. From our survey, we observe that the work in this field is in fast-paced development since the task of video description falls in the intersection of computer vision and natural language processing. But still, the work in the video description is far from saturation stage due to various challenges like the redundancy due to similar frames which affect the quality of visual features, the availability of dataset containing more diverse content and availability of an effective evaluation metric.
 
\end{abstract}

\ccsdesc[500]{Computing methodologies → Computer vision}
\ccsdesc[500]{Natural Language Generation}
\ccsdesc[500]{Machine Learning→ Supervised and Unsupervised Learning →Reinforcement Learning}
\keywords{captioning models, Datasets, Deep Learning Approaches, Evaluation metrics, Video description}

\maketitle

\section{Introduction}\label{sec:intro}
The rapid growth in the use of internet services in the past two decades has created a drastic increment in multimedia (especially video) data transmission and database size \cite{singh2019novel}. This rapid growth in the volume of multimedia data creates a necessity for a survey of effective techniques for the fast processing and storing of data \cite{abdulhussain2018methods}. In the current scenario, a video has become the most used datatype over the internet. For example, the number of video content uploaded daily over video sharing websites (such as YouTube, TikTok, etc.) as compared to earlier days is very high. This huge increment in the use of video data grasps the interest of the researcher towards proposing an automatic system that can describe the content, event, and action in the video in either a short textual form or generate a long informative description. Video is a rich source of information, it consists of multiple scenes, stories, action, events, text, and audio. The generation of a textual description of the video is a very challenging task. It involves not only just detecting the visual entities and their actions it aims to extract some logical or meaningful inference by establishing and predicting the relationships between them into a textual form. In the past decades, only the researchers of computer vision were solely working in this field. But, the recent success of Natural Language Processing (NLP) brings the researchers of both fields together to address the problem of the video description. Various special research works have been published for bridging the gap between the vision and languages through visual captioning.

\begin{figure}[!ht]
    \centering
    \includegraphics[scale=1]{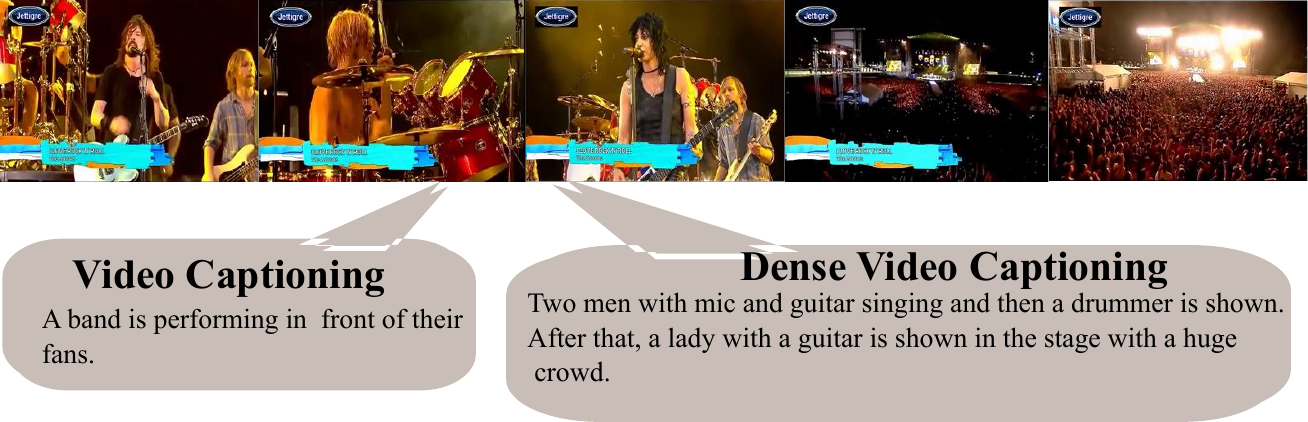} 
    \caption{Two types of natural language description: (1) Video captioning (short description) and (2) Dense video captioning (description in a paragraph).}
    \label{fig:dense_cap}
\end{figure}
There are various applications of video description such as storytelling, video sentiment analysis, video retrieval, and automatic tagging of videos based on the content. To provide more ease to human comfort, it can also help in making Artificial Intelligence (AI) devices capable of extracting some inference in making some decisions based on the text which is generated automatically based on the visual content. Both video description and image description are two main categories of visual description. The only difference is the nature of the content where the image has a static content which remains the same throughout the analysis whereas the video has a dynamic content which changes with time. This dynamic nature of the video makes the process of video description challenging. Based on the length of description generated for the video, it is divided into two categories viz,. video captioning and dense video captioning as shown in Figure \ref{fig:dense_cap}. From Figure \ref{fig:dense_cap}, we can see that the video captioning includes giving a concise idea of the overall visual events of the video whereas dense video caption involves the generation of a brief description of each and every scene in the video. Keywords like \textit{mic}, \textit{guitar}, \textit{singing}, \textit{drummer} and \textit{crowd} in example shown in Figure \ref{fig:dense_cap} are responsible for generating a long description.

\section{Structure of the Video}
\begin{figure}[!ht]
    \centering
    \includegraphics[width=6.8cm,height=6.8cm]{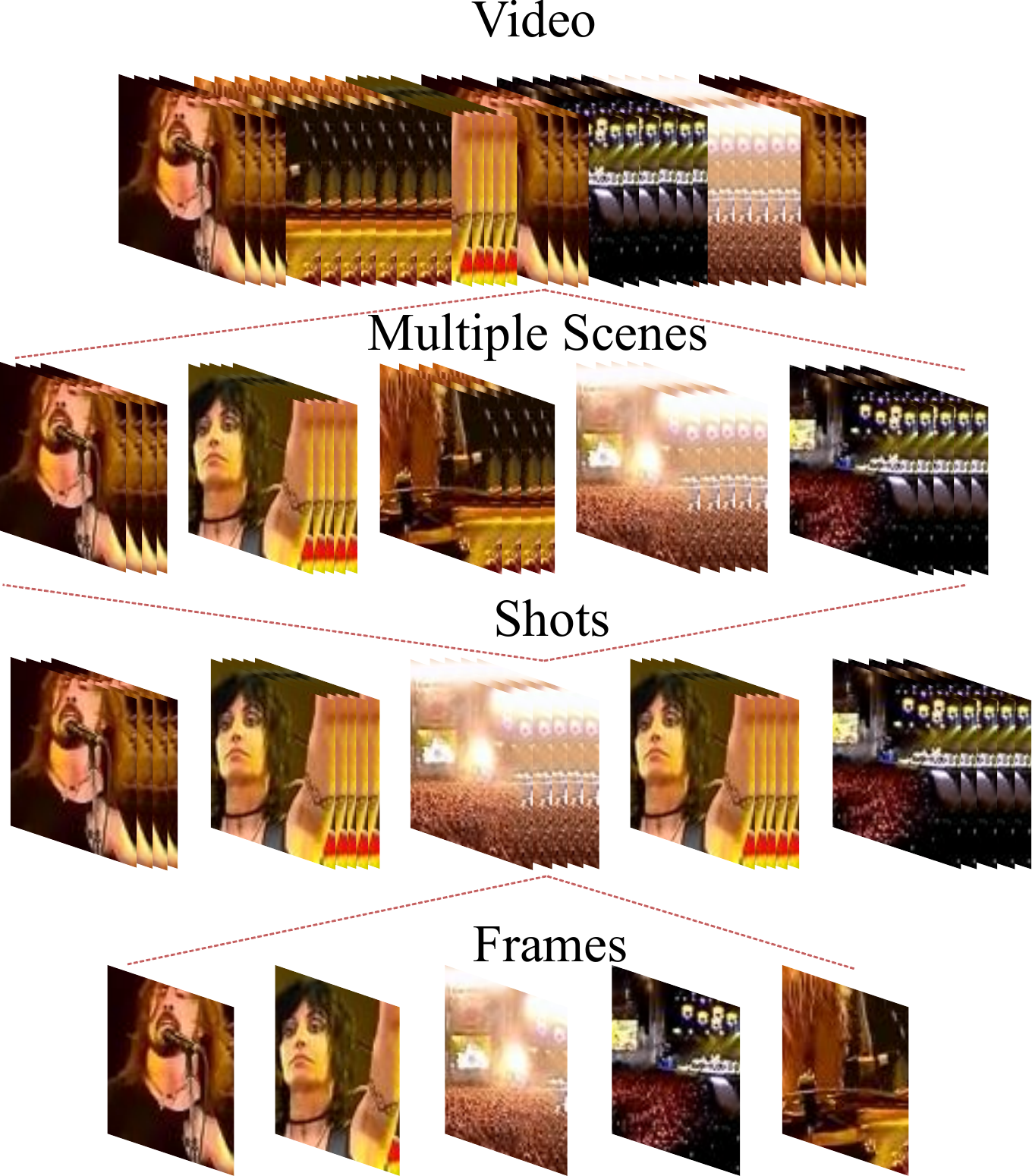} 
    \caption{Hierarchical structure of the video based on content of the frames.}
    \label{fig:video_hir}
\end{figure}

Before going in-depth about video description approaches and their performance, firstly we will explore the structure of the video. The performance of video description approaches totally depends on how effectively the visual features of the video are encoded. These encoded visual features should contain all the necessary characteristics required for classifying the event and action performed in the video. For encoding a video effectively, we should know about the hierarchy of the video, there are three levels of the video in which it can be divided which are named as scenes, shots, and frames. Based on the content, event, and action, a video can have multiple scenes. A scene consists of a collection of multiple interrelated contiguous shots captured from multiple angles. A scene is a collection of continuous similar frames taken uninterruptedly while a frame represents a single image. So, from Figure \ref{fig:video_hir}, we can see a video with a number of similar frames and the selection of the most informative frames is very crucial in the video description. That is why the selection of informative frames is the main focus of recent video description approaches \cite{chen2018less}.

\section{Video Description Approaches}
\begin{figure}[!ht]
    \centering
    \includegraphics[scale=0.5]{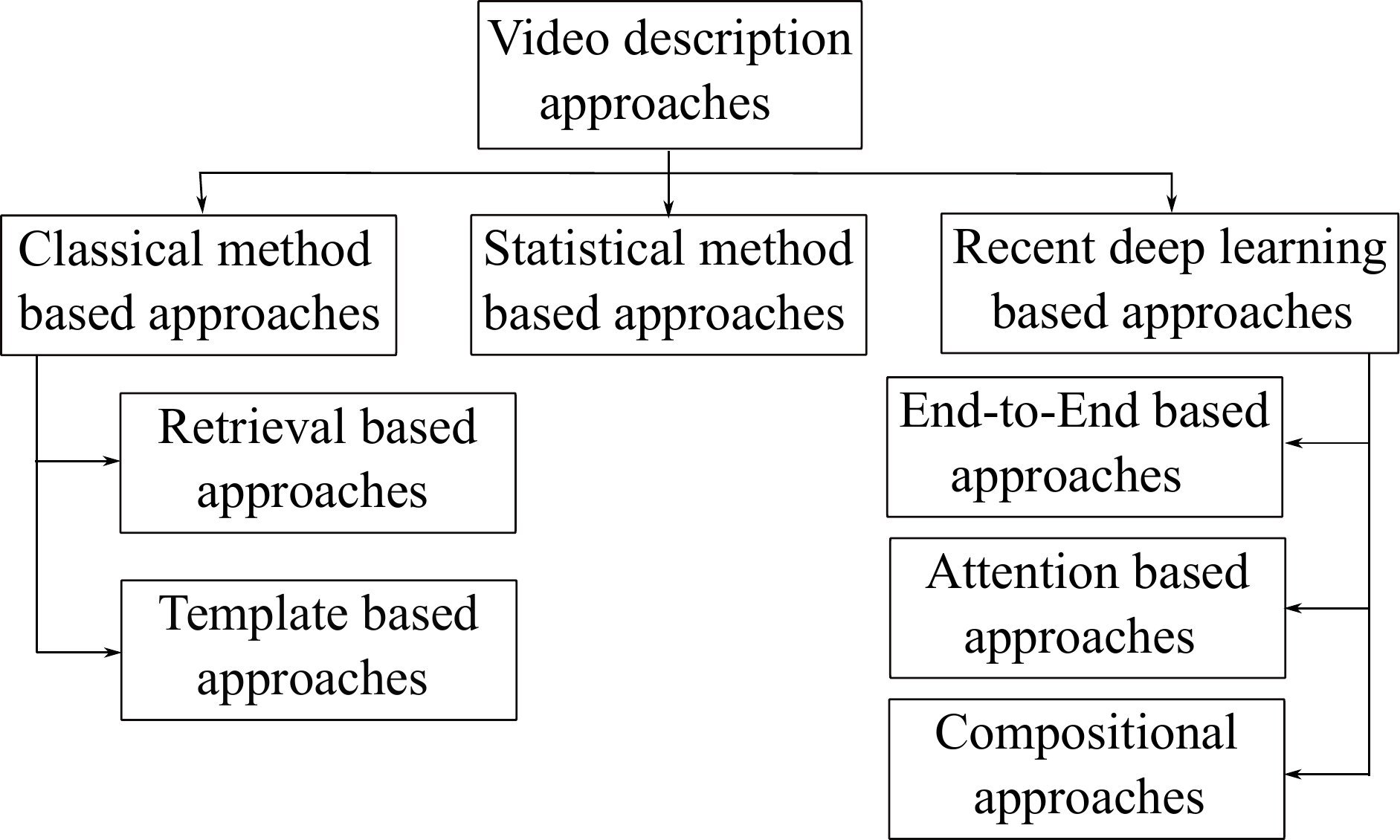} 
    \caption{Categorization of video description approaches into three phases.}
    \label{fig:video_description}
\end{figure}

\begin{figure}[!ht]
    \centering
    \includegraphics[scale=0.6]{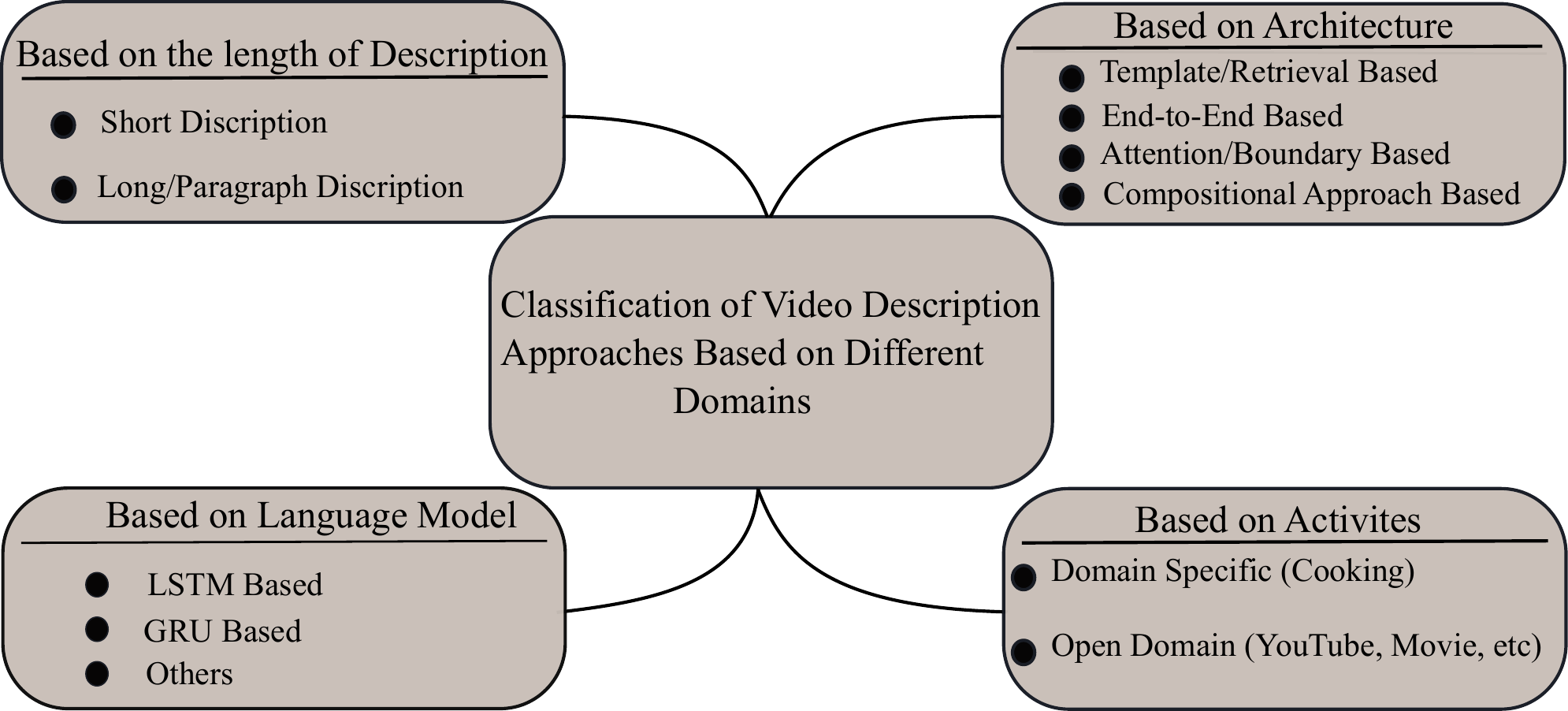}
    \caption{Video description approaches classification based on different domains.}
    \label{fig:vdc}
\end{figure}

In this section, a comprehensive study is performed on all the available main categories of video description approaches as shown in Figure \ref{fig:video_description}. The categorization of video description approaches can be done on various criteria such as based on the learning method used, type of output generated (a single sentence or descriptive), and type of model used which can be seen in the Figure \ref{fig:vdc}.

\subsection{Classical Video Description Approaches (CVDA)}
The main intuition of classical approaches for video description is basically based on either detecting the subject, object, and action in the video or retrieving the caption of the video having similar content. The early work in this field is started by \cite{kollnig1994association}. They proposed an approach for associating the motion verbs with vehicle movement. In this approach, they linked 64 motion verbs and generated a description of vehicle movement generated by tracking objects in real-world scenarios. Furthermore, \cite{brand1997inverse} extended the description generation process by generating semantic tags for the actions performed in the video. They proposed two approaches, one for extracting actions and keyframes from the video for storyboarding using "video gister" and in the second model, they proposed a framework that coupled the hidden Markov models to segment the video into actions. They considered this problem as 'Inverse Hollywood Problem' where the input is a video and output will be a script of action performed in the video along with keyframes containing these actions (such as in, out, add, move). The limitation of this approach is that the gister can only work for a video involving actions performed by only one human arm upon an object of arbitrary shape. The classical visual description approaches are two types named retrieval based video description approaches and template-based video description approaches
\cite{hossain2019comprehensive}. Most of the early-stage work in the video description is based on a template-based approach. In this approach firstly all related actions, events, and entities in a video are detected then using a fixed template a suitable description is generated. The rule-based systems are used for creating these templates and they are effective in a constrained environment (i.e., short clip, or the video with a limited number of actions and objects). The descriptions generated using template-based approaches are grammatically correct, but they lack in generating variable-length description \cite{hossain2019comprehensive}.

Basically, the main intuition behind all the classical video description approaches moves around the detection of the subject, object, and verb (SVO) in a video and fills them into a pre-defined template which consists of considering all the grammatical rules. Most of the traditional video description approaches can be considered as a two-stage process which includes \textit{visual attribute identification} (such as objects, the interaction between the objects) and \textit{natural language description generation}.

Understanding the human activities in a surveillance video is one of the most useful applications of video description generation. \cite{kojima2000generating} proposed an approach for understanding human behavior in a video and generating a natural language description. In the proposed approach, for estimating the human motion they have used the pose and position of the human head. They have estimated the pose of the human head from the sample images which shows high similarity. The pose of each sample image is previously known and for position estimation of the head they used projection transformation in the image plane. For generating a description of the human behavior into a textual form, a case grammar proposed in \cite{fillmore1967case} for machine translation is used. The drawback of the proposed approach is that it is only compatible for the videos having a single actor. It faces a lot of challenges if the number of actors will increase or if the scene is more complex.

In 2001, Ayers and Shah \cite{ayers2001monitoring} proposed a method for recognition of human actions and generating a textual description. Along with description generation, they have also generated a set of keyframes from the video which can be used for many applications such as content based video compression, indexing, and retrieval. The proposed approach only focuses on the actions which are performed inside a room such as enter, sit/stand, pick up objects, put down objects, open, close, using a computer, cabinet opening, leave, etc. To recognize these actions, the proposed model used prior knowledge about room layout. The components of the proposed model are skin detection based on color, tracking of multiple objects, scene change detection, and action recognition. The drawbacks of the proposed approach are confusing model when two-person get close and the performance of the model gets degraded if the skin area of the person gets occluded and if the quality of prior knowledge provided to the system is not as required.

Kojima and Tamura \cite{kojima2002natural} extended their previous work \cite{kojima2000generating} by recognizing the action performed by each body part in a textual form and integrating them into a final one description. For the estimation of human posture, three types of geometric information are used named as \textit{position of head}, \textit{direction of head} and \textit{position of hands}. In this approach, the pixel difference between the input and the background image is used for the detection of humans in the input frame and shape based matching with predefined objects is used for action recognition. After recognizing the actions performed by humans, they are associated with semantic primitives. Then, the whole body actions are integrated into a case frame using three types of integration rules. Finally, the description is generated using semantic rules and a word dictionary. Unlike previous work, in the proposed approach, the author tried to address the whole body action but still the model lacks when a video has multiple people acting independently.

Hakeem et. al. \cite{hakeem2004case} addressed the drawbacks of Kojima and Tamura \cite{kojima2002natural} by proposing three extensions to the case framework ($CASE^{E}$). They proposed a hierarchical CASE to address the issue of multiple dependent and independent agents in a video by using a case list of sub-events. For the proper understanding of the temporal occurrence of events, they included temporal logic based on interval algebra and they incorporated the casual relationship between the events because the occurrence of some sub-events depends on other events and some sub-events are independent.

Tena et. al. \cite{tena2007natural} tries to address the issue of generating a natural language description of human action in three languages (Catalan, English, and Spanish). The proposed framework has three subsystems named Vision Sub-system (VS) which include detection and tracking of objects in a video, a Conceptual Sub-system (CS) convert acquired structural knowledge from VS into logical knowledge for further manipulations and reasoning, and a Natural Language Subsystem (NS) for the generation of the description of human actions. The Natural Language Subsystem has three steps: Lexicalization, Discourse Representation, and Surface Realization. Lexicalization converts predicates generated from the CS into linguistic entities such as agents, events, and objects. Discourse Representation Structures (DRSs) overcomes the ambiguities in natural language terms by embedding semantic features using proper syntactical forms and in the surface realization stage morphological processes are applied over each word.

By considering retrieval of video based on the content and human-readable queries, Lee and Zhu \cite{lee2008save} proposed an approach named automatic semantic annotation of visual events (SAVE). This approach consists of three stages: image parsing, event inference, and text generation. Unlike Kojima et. al. \cite{kojima2002natural} where the case framework is used for the text generation process, grammar-based similar approach is used for annotating the scenes and events of the video. In the first stage, scene content is extracted using bottom-up image analysis. Then the extracted content from the first stage is passed to the event inference stage where Video Event Markup Language (VEML) is employed for representing semantic information. Then, finally the description is generated using Head-driven Phrase Structure Grammar (HPSG). Unlike, the previous approach, the generated description of the proposed approach provided more information regarding visual entities present in the scene and the relationship between them. For experimentation, they have focused on urban and maritime environments for getting a better description.

\begin{figure}[!ht]
    \centering
    \includegraphics[scale=0.5]{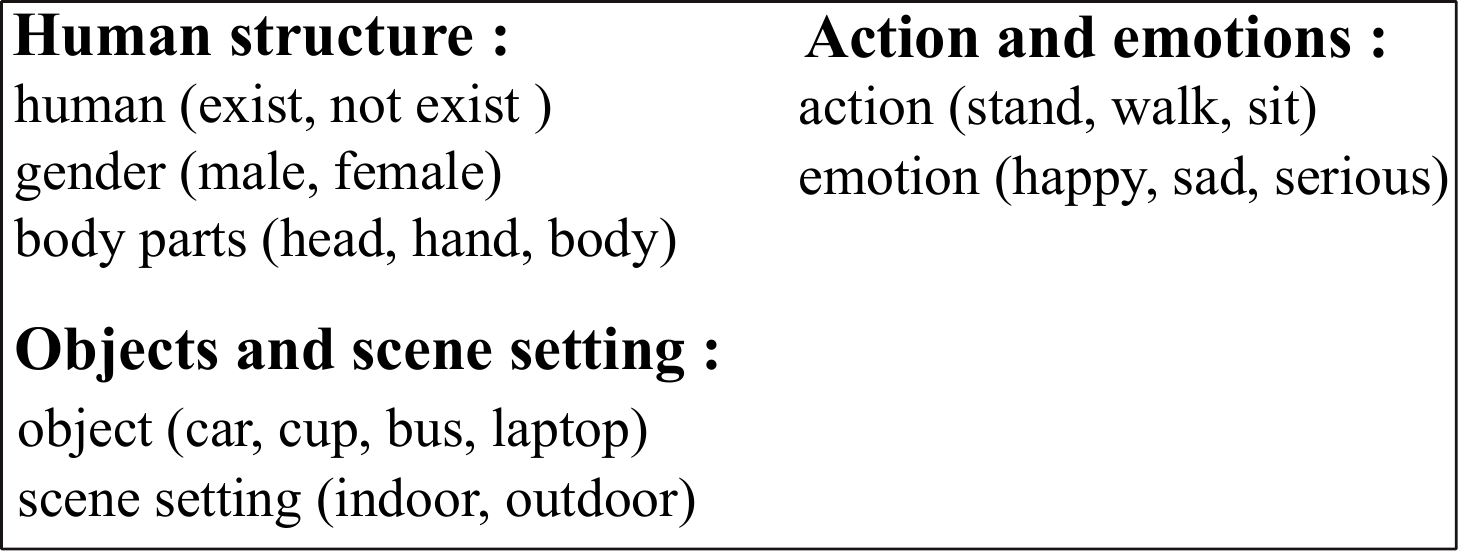}
    \caption{Sample predicates used for sentence generation.}
    \label{fig:paperkahn}
\end{figure}

Khan et. al. \cite{khan2011human} proposed an approach for generating the description of the actions performed by humans in a video. In the proposed approach, the gender of a human in a video is predicted with the help of face recognition, emotions, and actions performed by a human. For this task, they have used rushes videos proposed by TRECVid in 2007 and 2008 for video summarization tasks. This dataset contains 50 segments with one shot of length 5 to 20 seconds. Apart from reference annotation of 1 to 7 sentences for each segment which was provided by TRECVid, they have also used 20  human subjects short descriptions with multiple short sentences named as hand annotations. They employed a template-based approach for generating description. Figure \ref{fig:paperkahn} shows the sample predicates used for sentence generation. Since the proposed approach generates a description related to actions performed by a person so it can not generate a description in the absence of a person and also this approach can only identify pre-defined actions only.

Most of the approaches proposed till 2012 focus on the actions performed by a single person in the video. But, since an event in a video may include multiple actions performed by single or multiple persons to address this issue, Hanckmann et. al. \cite{hanckmann2012automated} proposed an approach by considering a wide range of events in a video. In the proposed approach, they have addressed 48 types of actions. This proposed approach has five main stages named visual processing, fusion engine, event description, action classifier and description generator. In this proposed approach, two types of action classifier are employed viz., discriminative bag-of-features classifier and generative rule-based classifier. The rule based system used in the proposed approach consists of 73 rules for 48 actions. The advantages of the proposed approach are that actions performed in a video are not restricted. In this approach, the DARPA dataset is used which has around 10 sentences as a ground truth. The limitation of the proposed approach is that the subjects and objects are categorized into four classes (person, car, bike, and other) due to which the classification accuracy of subject and objects is low.
    
Barbu et. al. \cite{barbu2012video} proposed an approach for generating a description for a short video in which they tried to address the shortcoming of earlier approaches such as generating a description for a video involving the interaction of more than one person. The proposed approach consists of three steps namely object detection \cite{felzenszwalb2010cascade}, object tracking \cite{shi1994good,tomasi1991detection} and dynamic programming. The object detection step includes the detection of objects in each frame. The number of objects detected per frame is limited to 12 for avoiding over detection and dynamic programming helps in the selection of the optimal set of detected objects. For generating labels for the video, Hidden Markov Model (HMM) is employed with a set of grammar rules and finally description for the video is generated by fitting these generated labels into a predefined template. The shortcoming of the proposed approach is that the model is trained on limited objects and action classes of the video of highly constrained domains.


\begin{table}[!ht]
    \caption{Summary of all the classical Video Description Approaches.}
    \label{tab:summCVDA}
    \centering
    \begin{tabular}{|c|p{5cm}|p{7cm}|}
    \hline
       \textbf{ Reference} &\textbf{ Domain} &\textbf{ Approach/Findings } \\\hline
       \cite{kollnig1994association}  & Actions associated with vehicle movement& Developed a system for describing vehicles movements into natural language  \\\hline
       \cite{brand1997inverse} &Actions performed by single person&Developed a 'video gister' for extracting actions and key frames from the video for story-boarding\\\hline
       \cite{kojima2000generating}& Human actions in a surveillance video&Proposed an approach to describe human actions in a surveillance video based on the pose and position of the human head\\\hline
       \cite{ayers2001monitoring}&Actions performed by a person inside a room&The proposed approach only focuses on the actions which are performed inside a room such a enter, sit/stand, pick up object, put down object, open, close, using a computer, cabinet opening, leave etc.\\\hline
       \cite{kojima2002natural}&Actions performed by single person & In this approach, pixel difference between the input and background image is used for the detection of human in the input frame and for action recognition shape based matching with pre-defined objects is used.\\\hline
        \cite{hakeem2004case}&Addressed the actions performed more than one person in a video&Proposed  hierarchical CASE to address the issue of multiple dependent and independent agents in a video\\\hline
        \cite{tena2007natural}&Human focused description generation&Attempted to address the issue of generating natural language description of human action in three languages (Catalan, English, and Spanish). The proposed framework have three subsystems named Vision Sub-system (VS), a Conceptual Sub-system (CS) and  a Natural Language Subsystem (NS). \\\hline
        \cite{lee2008save}&Videos of urban and maritime environments are selected for experimentation&Three stage based approach is employed namely: image parsing, event inference and text generation.\\\hline
        \cite{khan2011human}& Rushes videos proposed by TRECVid in 2007 and 2008& Face recognition, emotions and action performed by a human are predicted for generating a description of the action performed by human in the video using template based approach.\\\hline
        \cite{hanckmann2012automated}&Video of DARPA dataset are used for addressing multiple actions performed by multiple persons in a video.& Rule based approach is employed, it contains around 73 rules for 48 actions.\\\hline
        \cite{barbu2012video}&Human focused videos involving interaction of more than one person. &Predefined grammar rules and template based approach are employed for generating short description of the video.\\\hline
    \end{tabular}

\end{table}

In the classical method phase, the description is generated by detecting visual entities (such as an object, action, scene) in a video and fit them into fixed predefined templates. The rule-based systems are used for creating these templates and they are effective in a constrained environment (i.e., short clip, or the video with a limited number of actions and objects). The drawbacks of classical video description approaches are: (1) approaches proposed in this phase are complex in nature, (2) they are either focused on a single person in the video or only describe the motion of the vehicle in the traffic, (3) only limited actions (which are predefined) are addressed and, (4) unable to generate fluent and coherent long description. Table \ref{tab:summCVDA} reports the summary of all the discussed classical video description approaches with their domain and approach/findings of the proposed method.

\subsection{Statistical Method Based Approaches}
In the second phase of video description approaches to overcome the challenges faced by classical video description approaches in describing the open domain videos, some statistical-based approaches are proposed. These approaches are inspired by the success of statistical machine translation based approaches in the translation of languages. 

The early work in generating video descriptions using a machine learning-based approach is proposed by Rohrbach et. al. \cite{rohrbach2013translating}. In this work, they tried to address important research queries such as what is the best approach for the generation of linguistic description of visual information and which visual information is described by humans even though it is not present in the video, etc. The learning process for generating a natural language description is divided into two stages. In the first stage, inspired by the probabilistic model used for image description, an intermediate semantic representation is learned. Then in the second stage, generating a natural language description from an intermediate semantic representation is treated as a translation problem, and for that, the statistical machine translation \cite{koehn2007moses} is used. An aligned parallel corpus of videos, semantic representations, and reference descriptions are used for learning both the semantic representation and textual description. For the performance evaluation, the proposed method is compared with other classical methods as well as with a baseline method without using any intermediate semantic representation and language model.

The drawbacks of classical video description approaches such as complexity and scalability (when the model deals with the large vocabulary of open domain videos) are addressed by the statistical method based approaches to some extent by crafting the description generation process into two stages. First, visual entities and semantic labels recognition/generation phase and secondly, natural language generation phase. However, the division of the description generation process into stages creates a learning gap due to the inability of end-to-end training, which is addressed by the third phase of video description approaches, which include deep learning-based approaches. Further, in the next section, all the recent deep learning-based approaches are discussed with their advantages and disadvantages.

\subsection{Recent Deep Learning Based Approaches}

\begin{figure}[!ht]
    \centering
    \includegraphics[scale=1.2]{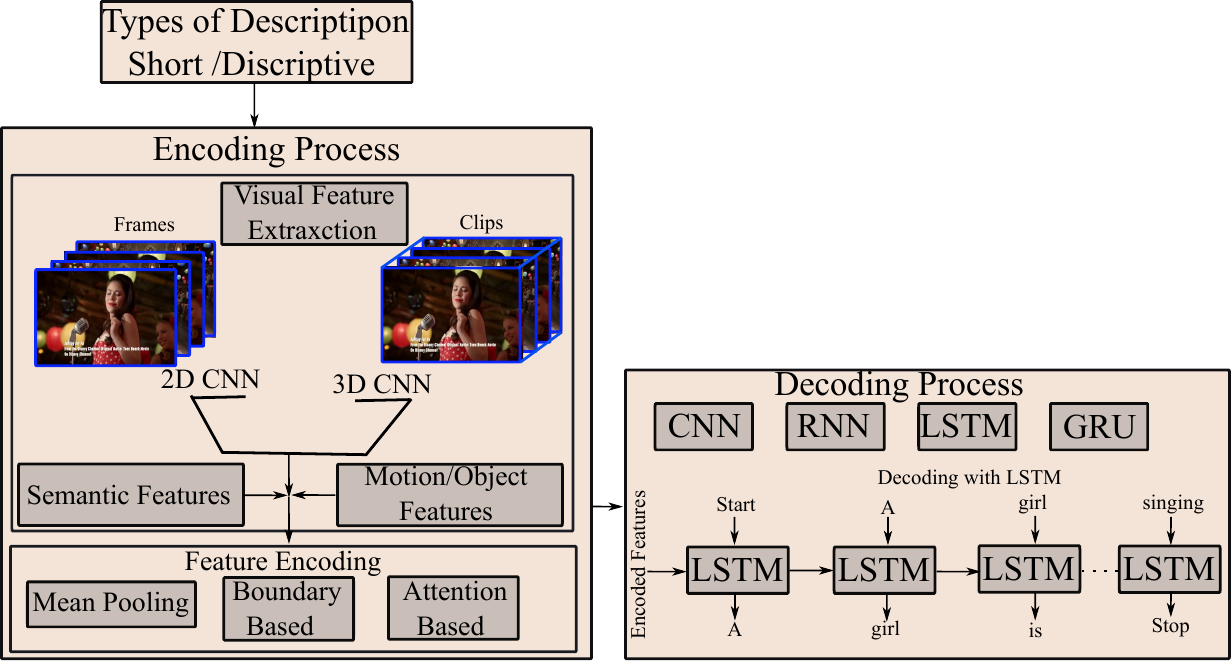}
    \caption{General block diagram of variants of deep learning based approaches.}
    \label{fig:bd}
\end{figure}

Most of the methods for generating text from a video are motivated by the image captioning methods. The recent success of deep learning in the CV and NLP field attract the interest of researchers for incorporating deep learning in the text generation process. The Convolutional Neural Network (CNN) and variants of the Recurrent Neural Network (RNN) such as Long Short Term Memory (LSTM) and Gated Recurrent Unit (GRU) have proved their ability in generating a video description. For a detailed analysis of the novel approaches proposed for the generation of video description, the proposed approaches are divided into three categories namely, End-to-End Based Approaches (E2E-BA), Temporal Boundary and Attention Based Approaches (TBaABA), and Compositional Approaches (CA). The Figure \ref{fig:bd} shows a general block  diagram of variants of deep learning based approaches which are mostly divided into two stages: encoding and decoding. 

\subsubsection{End-to-End Based Approaches (E2E-BA)}

For improving the quality of generated caption, several sequential CNN-LSTM, RNN-RNN, and reinforcement learning-based approaches (\cite{rohrbach2013translating}, \cite{dong2016early}) inspired by image captioning are proposed by using more appropriate features. Motivated by the success of CNN-LSTM based framework in image caption generation, \cite{venugopalan2014translating} proposed a similar framework for video captioning. To apply the method of image captioning on video \cite{venugopalan2014translating}, extracted feature from $fc_{7}$ layer (fully connected) for each frame in a video and then used a mean pooling approach to get a single feature vector of the whole video. These mean pooled features are supplied to an LSTM layer for the generation of textual description for a video. Since the video has a temporal structure of a sequence of frames and this method ignores the ordering of each frame.

Some researchers have tried to solve the problem of video captioning by using the novel Machine Translation (MT) approaches such as sequence to sequence (S2S) and transformer-based MT. \cite{venugopalan2015sequence} proposed an S2S method for visual description by using two LSTM layers. The first LSTM layer receives a sequence of frames as an input, it encodes them and then the hidden representation of the first layer is passed to the second layer. In the proposed approach different types of visual features are used such as convolutional features (using VGG16, AlexNet, and GoogleNet) and optical flow features which are passed to the input LSTM layer. During the training, the LSTM layers are unrolled for fixed 80-time steps, if the size of the video is less than 80-time step padding of zero is used for remaining input. For a video longer than 80-time steps is truncated to 80 frames.

In another approach, Venugopalan et. al. \cite{14} proposed an investigation on how linguistic knowledge can aid the video to text generation. In this approach, they amalgamated both neural language model and distributional semantic embedding for the generation of the text in three ways viz. early fusion, late fusion, and deep fusion. The experimental results of the proposed approach show that there is a significant improvement in the grammar of generated descriptions. The proposed model has experimented over both short and long description generation datasets MSVD (YouTube2text), MPII-MD, and M-VAD.

Krishna et. al. \cite{krishna2017dense} proposed a variant of an existing method for describing both short and long events. The proposed model is divided into two stages where in the first stage events in a video are detected and then in the next stage, a description is generated by capturing the dependencies between these events using both past and future events detail as a context. For event detection, an LSTM layer is used as a proposal module, when an event is detected the hidden state representation of that time step is used as visual features. In the proposed approach, two major challenges of a video description are addressed namely the duration of a particular event (beginning and ending of the event) and the relationship between the events in a video.

For generating an event-related text from a video, \cite{2} proposed a dual-stage based approach namely, event localization and paragraph generation for video description. In the first stage, localization of candidate event is performed using video action detection \cite{zhao2017temporal} and then the framework generates coherent and concise paragraph of description in the next stage where for avoiding the redundancy of due to similar events, a subset of unique events are selected using an event selection module. In this approach, all three modules including event localization, event selection and caption generation are trained separately.

\cite{daskalakis2018learning} proposed another method for learning spatio-temporal features for video captioning. This model contains three sub-networks. The first sub-network is used for extracting frame level features for each frame using the VGG16 model pre-trained on ImageNet Large Scale Visual Recognition Challenge (ILSVRC) dataset \cite{russakovsky2015imagenet}. The second sub-network contains three sub-modules for capturing spatial and temporal semantic information and global visual information of the video. The third sub-network is used for the generation of relevant captions. Each sub-network in the proposed approach aims to summarize the local and global dynamic concepts which help in achieving comparable results.

Aafaq et. al. \cite{aafaq2019spatio} observed that most of the video description framework focused on using a pre-trained CNN for visual feature encoding. Aafaq et. al. \cite{aafaq2019spatio}  argue that the encoding of visual features effectively plays a major role in generating a qualitative natural language description of the video. In the proposed work using Gated Recurrent Units (GRUs), an effective visual features encoding method is proposed for generating semantically rich descriptions of the video. In this method, Short Fourier Transform is applied to CNN features in a hierarchical manner for embedding temporal dynamic to visual features. The proposed approach also employs an object detector for getting more high-level semantic information. Finally, all the representations are mapped to a common subspace and passed to a recurrent layer of GRU for sequence modeling.

Nabati and Behrad \cite{nabati2020multi} proposed a video captioning framework by employing a multi-stage refining algorithm and content-oriented beam search. The proposed approach includes three stages namely feature extraction, content-oriented beam search and sentence refining. In the first stage, using deep CNN, visual features are extracted along with the object detector algorithm \cite{redmon2018yolov3} for recognizing the objects in a frame which are used as semantic attributes. In the content-oriented beam search encoding-decoding process, at each time step k, most probable words are selected using a beam search and finally, each of them will generate a probable caption for an input video. In this stage, a structural dictionary \cite{davies2010corpus} is also employed for generating a structurally correct sentence. Finally, the refining stage performs two types of refining namely structural filtering and content-based filtering. As the name suggests, structural filtering removes the sentences which have some grammatical deficiency or repetition of words and content-based filtering removes the sentence that has less semantic content than the actual semantic content present in the video. In the proposed approach similar to \cite{venugopalan2015sequence}, only 80 frames are used for reducing the computational time.   

Singh et. al. \cite{singh2020nits} proposed a video captioning framework in the VATEX challenge using two parallel LSTMs. The way of fusing visual representation with an embedded representation of the reference caption is different for both LSTM. For the first LSTM, the initial states are initialized with visual representation and for the second LSTM, the visual representation is concatenated with embedded representation parallelly at each time step. Finally, the most probable word at a particular time step is predicted after performing a dot product between the output of both LSTMs.

Zhu et. al. \cite{zhu2019multi} proposed a framework in the VATEX-2020 video captioning challenge by improving their earlier model. In the proposed work, X-Linear model is employed which performs better than the top-down model \cite{zhu2019multi}. To improve the model's understanding, multiple features like the combination of appearance features and motion features are employed. In the proposed work, it is observed that the hybrid reward strategy which is a combination of different automatic metric scores yields better results. The proposed framework outperforms other approaches in the challenge.

For encoding the sequential frames into a spatiotemporal representation, \cite{wang2020sequence} proposed a `sequence in sequence framework' (SeqInSeq) for generating a word at each time step. In this framework, sequential frames are encoded at each time step for uttering a most related word in the sentence. For the aggregation of the visual features at each time step, a ``Real-Time Encoder" (RTM) is proposed that learns to extract spatiotemporal features at each time step. It uses history information of the previous time step for ensuring more informative visual features for the generation of the most suitable word in the caption.

On observing that most of the existing video description models ignore the interaction between the objects.  Motivated by this, Zhang et. al. \cite{zhang2020object} proposed a novel video description model and a new training strategy. In this work, an encoder based on a learnable object-relational graph (ORG) is proposed for the full exploration of spatial and temporal relationships between the objects. The ORG is able to extract effective interaction features for enhancing visual representation. A teacher-recommended learning (TRL) approach is also proposed for overcoming the shortcomings of teacher enforced learning. TRL makes full use of the external language model (ELM) for leveraging the video description model with linguistic knowledge and generates more similar word proposals for long-tailed problems. The proposed model has been experimented on MSVD, MSR-VTT and VATEX datasets.



\subsubsection{Temporal Boundary and Attention Based Approaches (TBaABA)}

A video is a collection of a large number of continuous frames in which the similar frames are repeated several times due to which the problem of frame redundancy occurs and affects the process of generating video descriptions. The above work ignores the problem of frame redundancy in a video and gives the same weightage to each frame during the encoding phase due to which the final representation consists of large redundant information and lacks informative features that are required for generating relevant captions. To address the problem of frame redundancy and temporal dependencies between the frames, many boundary aware video description and temporal attention-based approaches have been proposed using temporally aware long short term memory framework which can detect scene change in the video. Since the attention based approach has shown promising results in the image captioning domain motivates the researchers to apply temporal attention in a video.

Using both the local and global temporal structure of the video, \cite{12} proposed an approach for the generation of text from the video. In this paper, firstly special temporal 3D-CNN is used for generating the representation of the short temporal dynamics. Second, they also proposed a temporal attention mechanism that helps to go beyond temporal modeling and select the most relevant segment for the given video. The 3D-CNN model is trained on video action recognition tasks to generate a representation which contains more human motion and behavior features. The shortcoming of this approach is that they have considered only the first $240$ frames from each video for reducing the memory and computational complexity. But, there is a possibility of missing some important information that is present in the remaining frames.

Using temporal video segmentation, Shin et al. \cite{Shinet} proposed a method for the generation of the video description. In this method, the temporal video segmentation is performed by employing sliding windows of length 30, 60, 90, and 120 frames and non-maximum suppression for removing redundant frames. The proposed approach is based on the concept of boundary detection which is intuitive for video based tasks but it has few drawbacks such as the sliding windows based approach does not guarantee the effective temporal segmentation of the video, it may miss some boundaries in the video.

Chen et. al. \cite{chen2018less} proposed a method for video captioning based on the assumption that selecting informative frames from the video is more effective in video description generation and in reducing computational time. For the selection of informative frames from the video, a plug and play PickNet is proposed which can be used in other video-related applications for selecting the subset of most informative frames from the video and removing redundancy due to similar frames. The experimental results show that the proposed video captioning model uses 6-8 frames for the generation of the description of the video. 

Most of the open domain videos consist of several shots and each shot is nearly related to each other. To understand the temporal structure of the video and minimize the redundancy of frames, the model should be able to detect the temporal boundaries in a video effectively. Some of the approaches are as follow:
 
Considering the discontinuity in the content of the video, Baraldi et. al. \cite{Baraldiet} proposed a novel LSTM cell for detecting the temporal boundaries in a video and generates a visual feature vector for the whole video which is passed to another LSTM layer and final captions are generated using GRU. The intuition of detecting boundaries is to ensure that the visual features of the current segment are not influenced by the segment which is already seen before and generate a hierarchical visual representation. The proposed boundary detection module can be applied to other video related problems such as video classification and action detection.

Considering the hierarchical structure of the video, \cite{PanPingbo} proposed an approach named Hierarchical Recurrent Neural Encoder (HRNE) to exploit temporal information in a video. The proposed approach has three main contributions. First, effectively handling temporal information. Secondly, computationally efficient, and third, uncover the temporal transition between the frames as well as between the segments. The limitation of the proposed approach is that the length of all videos is restricted to 160 frames. If the number of frames is more than {160} then they are dropped. If the video does not have enough frames, it is padded by zero frames.

Chen et. al.\cite{chen2019boundary} modifies the traditional encoding and decoding framework for video captioning by using the hierarchical encoder proposed by \cite{Baraldiet} and temporal attention mechanism proposed by \cite{12} for automatically selecting a relevant frame from the input video. In this approach, GRU is used as a decoder for generating the description of the video. In the proposed approach, the experimentation is carried out on two open domain datasets.

By combining the semantic information of the video with an audio-augmented feature, \cite{xu2019semantic} proposed an approach for generating a short video description. In the proposed approach, they proposed a semantic-filtered LSTM video encoder and soft-split-aware-gated multilayer LSTM encoder for extracting audio-augmented features and semantic features respectively. In the proposed approach, the a novel LSTM based architecture is also used for the detection of boundaries the video. The proposed approach has been experimented on three open domain video datasets viz., MSR-VTT, YouTube2Text and M-VAD.

Huanhou Xiao et. al \cite{XiaoHuanhou} proposed a method for video understanding using hierarchical LSTM and dual staged loss for converting the video into a sentence. In this approach, convolutional architecture is employed for extracting the fragment level features and then they are passed to a three-layer LSTM network where the first LSTM layer is used for encoding visual feature and the next two LSTM layers are employed for generating a description where local loss (Loss1) is for optimizing the second LSTM layer and global loss (Loss2) is used at the end for ensuring encoder-decoder traning process. To avoid the use of visual attention for non-visual words, a blank feature vector (contains all zeros) is appended with the encoded visual features. The objective function used in the proposed approach is shown in Equation \ref{eq:1} where Loss1 and Loss2 are cross-entropy losses.

\begin{equation}\label{eq:1}
    L = (1- \delta)Loss1 +Loss2
\end{equation}

A method for video captioning by exploiting the combination of both semantic and temporal attention is proposed in \cite{gao2019fused}. The main focus of the proposed approach was on designing an efficient approach that can make use of semantic concepts of the video for improving the quality of generated captions. The framework consists of two sub-networks viz., semantic concept learning network which is used for extracting semantic features and hierarchical decoder network which is based on Gated Recurrent Unit (GRU).

The success of transformer in MT inspired \cite{10} to propose a method for video captioning using a transformer-based model. They used C3D and I3D for extracting the spatiotemporal features for captioning. In the proposed approach, the $fc_6$ layer of C3D is used for feature extraction. The frames are sampled in the interval of 20 and I3D features are extracted for a batch of 8 frames and Principal Component Analysis (PCA) is used for the reduction of feature dimension to 512. The drawbacks of the proposed approach are that the proposed model is unable to recognize activities efficiently and can not generate coherent visual descriptions after the successful recognition of activities and nouns in the video.

For generating an accurate and temporally related description of the video content, it is necessary that the model should learn both temporal dynamics and visual concept inspired by the concept Li et. al. \cite{li2018multimodal} proposed for a multimodal video captioning framework using an attention mechanism and memory network. In this paper, soft-attention \cite{lu2017knowing} and a memory module is employed for selecting relevant input features since words like 'of' and 'the' does not require any visual feature since the generation of these words mostly depends on the previously generated word. The proposed model has been experimented over two open domain datasets MSR-VTT and MSVD and the model shows comparable results.
 
Olivastri et. al. \cite{9022152} proposed an approach for video description with two-stage training. In the first stage of training, the proposed architecture is initialized using pre-trained encoders and decoders and then fine-tuning in the second stage of the training. The whole network is trained end-to-end for learning the most relevant visual features. In the proposed approach, Inception-ResNet-v2 and GoogLeNet are used as encoders and a soft-attention-based LSTM (SA-LSTM) as a decoder. Both encoder and decoder are optimized simultaneously in an end-to-end fashion. From the experimentation, it is observed that the end-to-end training of the model performs better than other conventional disjoint training processes. The proposed work experiments are on two open domain datasets, viz., $MSR-VTT$ and $MSVD$.
 
Xu et. al \cite{xu2020video} proposed a channel and temporal-spatial attention mechanism on observing the performance of attention mechanism in generating natural language description of visual entities. The proposed attention mechanism helps the model to benefit from visual features along with maintaining the consistency between visual features and sentence descriptions. The experimental work of the proposed work shows that the proposed attention mechanism is successful in generating the words related to visual entities in the video.
 

Since most of the video description approaches optimize the loss of the model by considering all samples in a batch and ignore the learning situations.  Motivated by this, Xiao and Shi \cite{xiao2020video} proposed a text based dynamic attention framework (TDAM). Unlike other previous attention based approaches in this framework, dynamic attention is applied on the generated worlds for improving semantic context information.  The visual attention is also employed along with dynamic textual attention for selecting the most relevant word. Further, in this framework for training the model, two approaches are adopted:  ``starting from scratch" which optimizes the model using all samples and ``looking for gap" in which the model is trained for the samples with poor control. The proposed framework is able to achieve comparable results with state-of-the-art approaches on MSR-VTT and MSVD datasets. 

Nabati and Behrad \cite{nabati2020video} proposed a novel boosted and parallel framework for generating a natural language description of the video. The proposed framework includes two LSTM layers and a word selection module. The visual features of the frames which are extracted using a pre-trained CNN are encoded using a first LSTM layer. The second LSTM layer is known as Boosted and Parallel LSTM (BP-LSTM). This layer includes several LSTMs for decoding and enhancing the generated description in parallel and boosting framework. For training the network, AdaBoost algorithm is employed in a training phase. In the testing phase, the output of BP-LSTMs are combined. Then, the most probable word is selected using the Maximum Probability Word Selection (MPWS) module. 

Considering the application of video understanding in the field of video surveillance, Sah et. al \cite{sah2020understanding} proposed a video description framework. In the proposed work, a novel attention mechanism is used with the hierarchical architecture at both global and local levels. A recurrent model with soft attention is used with the dynamically detected abrupt (cut) transition in a video for an effective multi-stream hierarchical video description model. The proposed end-to-end video description model is also able to spot the temporal location of the actions and objects in the video. In the experimentation, on observing the timing and fidelity of the proposed video description model, the authors have claimed that the proposed model is suitable for the automation of surveillance systems. In support of this claim, a real-time analysis of the proposed video description approach is also performed. In the proposed work, a single system is sufficient for generating a natural language description which shows the computational efficiency of the proposed framework by altering the frame rate for 100 simultaneous cameras.   

Liu et. al. \cite{liu2020sibnet} proposed a novel visual information encoder named Sibling Convolutional Encoder (SibNet). The proposed SibNet has two branches viz., content branch and semantic branch. Both the branches encode video collaboratively. The content branch is used for encoding the visual information in the video using an autoencoder and the semantic branch uses a visual-semantic joint embedding which encodes semantic information. Then, the output from both branches is combined and it is passed to a decoder RNN using a soft-attention for generating a natural language description. In this architecture, a new loss function is proposed for jointly training both the encoder and decoder. The loss function includes three-loss terms: (1) content loss, (2) semantic loss and (3) decoder loss. The Equation \ref{eq:ml} shows the proposed loss function where $L_d$ is decoder loss, $L_c$ is visual content loss and $L_s$ semantic loss. $\alpha$ and $\beta$ are two scalars for regulating effect of content loss and semantic loss respectively. The values of $\alpha$ and $\beta$ are set to $0.4$ and 1 respectively by cross-validation.

\begin{equation}\label{eq:ml}
    L = L_d + \alpha L_c + \beta L_s
\end{equation}


\subsubsection{Compositional Approaches} 
Unlike E2E and TBaA based approaches, compositional approaches include the amalgamation of reinforcement learning with visual and semantic attention. Since, in the sequential video description models, the cross-entropy loss is optimized at word level during training but not at the sentence level. Motivated by this, Pasunuru and Bansal \cite{pasunuru2017reinforced} proposed a reinforcement learning-based approach using policy gradient and mixed loss. In this approach, the sentence level metrics is directly optimized which shows improvement over baseline approaches in terms of human evaluation and automatic metrics. Further, a novel entailment-enhanced reward (CIDEnt) is proposed which is an improvement over CIDEr which is a phrase matching metrics. In this approach, the visual attention is used for getting a visual context vector and a mixed loss is used which consists of cross-entropy loss (XE) and reinforcement learning loss (RL). The final mathematical expression for mixed loss is shown in Equation \ref{eq:reinforce1} where $\gamma$ is the parameter to be tuned to balance both losses.

\begin{equation}\label{eq:reinforce1}
    L_{MIXED} = (1-\gamma)L_{XE} + \gamma L_{RL}
\end{equation}

Wang et. al. \cite{wang2018video} proposed a hierarchical reinforcement learning approach for generating a video description. In this approach, a high-level manager is employed for formulating sub-goals. Low-level worker module is employed by selecting primitive actions at each time-step for the recognition of actions and to fulfill sub-goals. For determining whether the goals are achieved or not, an internal critic is used. For the selection of the relevant features, the manager and worker module employ an attention module. In this work, the authors have also introduced a novel video description dataset named Charades Captions\footnote{Charades Captions which was obtained by preprocessing the raw Charades dataset \cite{sigurdsson2016hollywood}.}. The proposed approach is able to generate both single and multiple sentence descriptions which is an advantage of the proposed approach over \cite{krishna2017dense} which does not address the scenario of single sentence generation.

Li and Gong \cite{li2019end} addressed the issue of end-to-end (E2E) training of video description models which is challenging due to memory consumption and data-hungry nature of deep recurrent networks. In this method, a multitask reinforcement learning approach is proposed for E2E training of the video captioning model. The proposed work mine and construct as many as attributes from the reference caption to make the learning process more effective, in the experimentation, it is claimed that the proposed E2E multitask reinforcement learning approach is successful in keeping the model-free from overfitting. The experimental results show that the model is able to outperform baseline models for MSR-VTT and MSVD datasets.

Recently, Wei et. al. \cite{wei2020exploiting} proposed another method for video captioning by employing reinforcement learning-based and temporal attention. For better use of temporal events, an adaptive sliding window is used. To predict the size of the sliding window, a reinforcement learning algorithm is proposed. For generating more relevant words, temporal attention is employed to selectively focus on specific temporal information. In this approach, reward (CIDEr score is used as a reward) based loss function is employed and the gradient is approximated using a single Monte-Carlo sample as done by Rennie et al. \cite{rennie2017self}. For extracting the visual features, K frames are used for each video and frame-level static features and motions features are extracted respectively for each video using ResNet152 model \cite{he2016deep} and 3D CNN model \cite{hara2018can}.

\begin{table}[!ht] 
\caption{Summary of  compositional approaches and their performance on MSVD, MSR-VTT, and Charades datasets}
    \label{tab:performcomp}
    \centering
   \begin{tabular}{|c|c|c|c|c|c|c|c|c|c|}
    \hline
\multirow{2}{*}{\textbf{Methods}} & \multicolumn{3}{c|}{\textbf{MSVD}}&\multicolumn{3}{c|}{\textbf{MSR-VTT}}&\multicolumn{3}{c|}{\textbf{Charades}} \\\cline{2-10}
&\textbf{BLEU-4}&\textbf{METEOR}&\textbf{CIDEr}&\textbf{BLEU-4}&\textbf{METEOR}&\textbf{CIDEr}&\textbf{BLEU-4}&\textbf{METEOR}&\textbf{CIDEr}\\ \hline
\cite{pasunuru2017reinforced}&54.4&34.9&88.6&40.5&28.4&51.7&&&\\\hline  \cite{wang2018video}&-&-&-&41.3&28.7&48.0&18.8&19.5&23.2\\\hline
\cite{li2019end} &50.3&34.1&87.5&40.4&27.0&48.3&-&-&-\\\hline
  \cite{wei2020exploiting}&46.8&34.4&85.7&38.5&26.9&43.7&12.7&17.2&21.6\\\hline
\end{tabular}
\end{table}

The compositional approaches addressed some main issues of the video description process such as learning in limited data, memory consumption in E2E learning and training deep reinforcement models based on reward functions such as BLEU-4 and CIDEr score. Since the reinforcement learning based approaches discussed above just have a comparison with baseline video captioning models with respect to datasets not among themselves. So, the Table \ref{tab:performcomp} reports the performance of each method discussed above to make a fair quantitative comparison among all the proposed compositional approaches.

\section{Datasets }\label{datas}
For the success of a particular domain, the availability of a standard dataset plays a crucial role because it contains most of the challenges of the particular domain. The datasets available for the video description differ in the various perspectives such as the number of reference captions, number of videos in the dataset, the domain of the videos and size of the videos. Based on the domains, the datasets can be classified into open domain video and domain-specific videos. The open domain video consists of videos from YouTube, Vines and Movies, etc., while most of the domain-specific datasets include videos from the cooking domain. Further, for simplicity, we have divided the section into a domain-specific dataset and open domain datasets.

\subsection{Domain Specific Datasets}
\subsubsection{MP-II Cooking (2012)}
In 2012, Rohrbach et. al., \cite{rohrbach2012database} put a stepping stone in the field of video description by proposing a fine grain domain specific dataset containing the description of 65 cooking activities (like cut, wash, chopping, or spice) which are performed by 12 participants preparing different dishes. Along with textual descriptions, the authors have also generated annotations for different body positions. For annotation, a two-stage revision process is used for at least 30 frames (1 second). A background activity is generated automatically for the detection task when an activity that is not annotated is performed or when a person is not present in the segment. The dataset includes 44 videos with 5,609 annotations. The average length of each clip is approximately 600 seconds. The main drawback of the MP-II dataset is that all videos are captured using a fixed camera inside the same kitchen with alike background \cite{aafaq2019video}. 

\subsubsection{YouCook (2013)}
\begin{figure}[!ht]
    \centering
    $\begin{array}{c}
         \includegraphics[scale = 0.3]{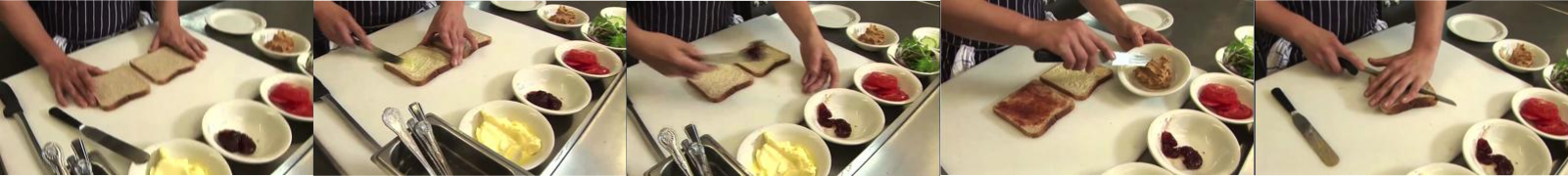}  \\
       \textbf{ Ref1:}\text{ In this video, the man cook some food on the direct heating pan... \ \ \ \ \ \ \ \ \ \ \ \ \ \ \ \ \ \ \ \ \ \ \ \ \ \ \  \ \ \ \ \ \ \ \ \ \  } \\
        \textbf{Ref2:}\text{ Two pieces of brown bread are taken and butter is applied on one slice... \ \ \ \ \ \ \ \ \ \ \ \ \ \ \ \ \ \ \ \ \ \ \ \ \ }\\
       \textbf{ Ref3:} \text{ In this video, a cook kept two pieces of bread over the cutting board and spreads butter...}\\
    \end{array}$
    \caption{Sample annotation for YouCook dataset.}
    \label{fig:Youcook}
\end{figure}

Das et. al., \cite{das2013thousand} proposed another domain-specific cooking dataset named YouCook\footnote{Available at \url{www.cse.buffalo.edu/~jcorso/r/youcook}} made up of 88 videos clips downloaded from YouTube. This dataset is considered more challenging than the MP-II dataset. Because in this dataset, most of the videos are captured with varying kitchen environments and also incorporating different camera changes. This dataset focuses on six different cooking styles. For employing machine learning approaches, the training set contains 49 videos and 39 videos for the test set along with their corresponding annotations. To generate a multiple natural language description for each video, Amazon Mechanical Turk (AMT) is employed. The annotators are instructed to generate at least three sentences for each video with minimum of 15 words. Figure \ref{fig:Youcook} shows the sample annotation for Youcook dataset.

\subsubsection{Textually Annotated Cooking Scenes (TACoS, 2013)}
\begin{figure}[!ht]
    \centering
    $\begin{array}{cccccclll}
    \hline 
\text{\textbf{Sample frame}} &\text{\textbf{Start}}& \text{\textbf{End}}&\text{\textbf{Action}}&\text{\textbf{Participants}}&\text{\textbf{NL Seq 1}}&\text{\textbf{NL Seq 2}}&\text{\textbf{NL Seq 3}}\\
    \hline\\
 \multirow{4}{*}{\includegraphics[scale=0.1]{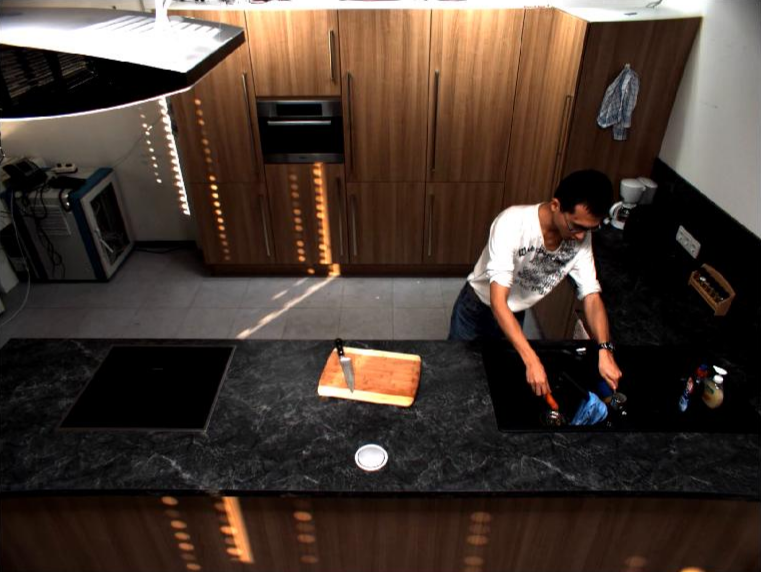} }&{\text{743}}&{\text{911}}&{\text{wash}}&{\text{hand, carrot}}&{\text{He washed carrot.}}&{\text{The person rinses }}&{\text{ He rinses }}\\
 &&&&&&\text{the carrot.}&\text{the carrot}\\
&&&&&&\text{ faucet.}&\text{from the}\\
  &&&&&&&\\
  &&&&&&&\\
  \hline\\
  \multirow{3}{*}{\includegraphics[scale=0.1]{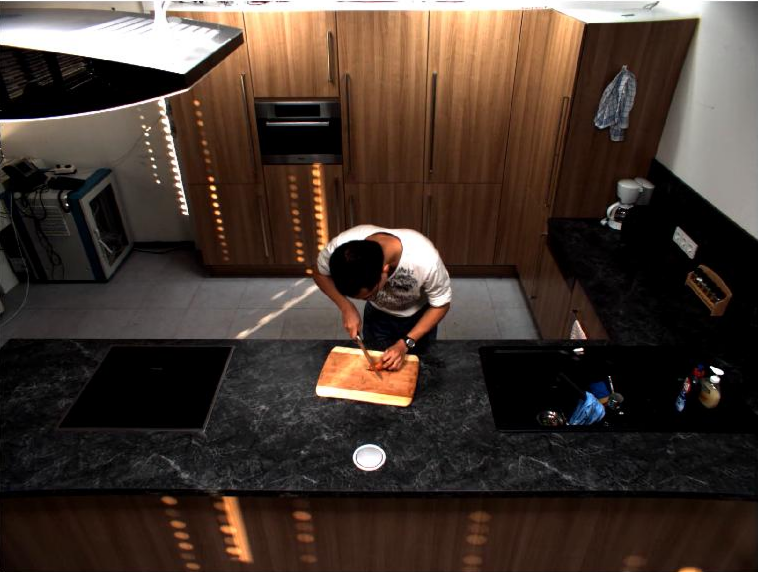}}&\text{982}&\text{1090}&\text{cut}&\text{knife,  carrot,}&\text{He cut off ends}&\text{The person cuts}&\text{He cuts off} \\
  &&&&\text{cutting board}&\text{of carrots.}&\text{ off the ends }&\text{the two}\\
  &&&&&&\text{of the carrot.}&\text{edges.}\\
  &&&&&&&\\
  &&&&&&&\\
  \hline\\
    \end{array}$
    \caption{Sample annotation for TACoS dataset.}
    \label{fig:Tacos}
\end{figure}

This dataset consists of videos of MP-II composites \cite{rohrbach2012script} dataset. The MP-II composites dataset includes 41 cooking activities in around 212 high-resolution videos. The length of the videos in this dataset varies from 1-23 minutes with an average length of 4.5 minutes. In this dataset, if a video contains meaningful cooking related action or movement of more than one body part ( such change in position or state of the object) then and only the annotation is provided for the segment. After refining the MP-II composites dataset, only 26 fine grained cooking activities in 127 videos are retained in the TACoS dataset\footnote{Available at \url{http://www.coli.uni-saarland.de/projects/smile/page.php?id=tacos}} \cite{tacos:regnerietal}. The TACoS dataset includes only those videos which have the activity of mixing ingredients and tasks having at least 4 videos in the dataset. In this dataset, each video has a corresponding 20 different reference descriptions. It includes 11,796 different textual descriptions including the narration of 17,334 actions. The size of the vocabulary is 28,292 tokens. The add-on of the TACoS dataset over the MP-II composites dataset is that it also provided a high quality alignment between sentence and video segment with their associated starting and ending time stamp. Still, this dataset also inherits the drawback of the MP-II cooking dataset which is having a static kitchen environment with a fixed camera that they have tried to address in the refinement phase of the videos. Figure \ref{fig:Tacos} shows the sample annotation for TACoS dataset.

\subsubsection{TACoS-MultiLevel (2014)} 
\begin{figure}[!ht]
    \centering
    $\begin{array}{c}
     \includegraphics[scale = 0.14]{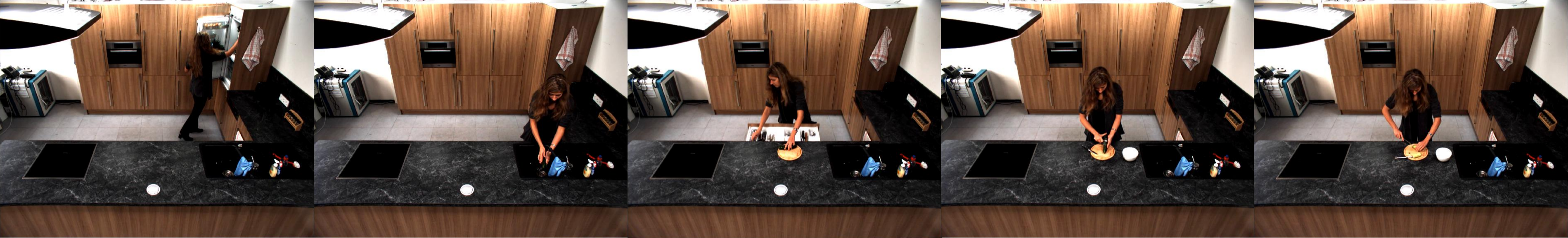}   \\
      {\text{ \textbf{Detailed:} A woman turned on stove. Then, she took  out a cucumber from the fridge. She washed the cucumber in \ \ }}\\
      \text{ the  sink. She took out a cutting board and knife. She took out a plate  from the drawer. She got out a plate. Next, she} \\
     \text{ took out a peeler from the drawer. She peeled  the skin off of the cucumber. She threw away the peels into the  waste}\\
      \text{bin. The woman sliced the cucumber on the cutting board. In  the end, she threw away the peels into the waste bin.}\\
    \text{\textbf{ Short:} A woman took out a cucumber from the refrigerator. Then, she peeled the cucumber. Finally, she sliced the \ \ }\\
    \text{cucumber on the cutting board. \ \ \ \ \ \ \ \ \ \  \ \ \ \ \ \ \ \  \ \ \ \ \ \ \ \ \ \ \ \ \ \ \ \ \ \ \ \ \ \ \ \ \ \ \ \ \ \ \  \ \ \ \ \ \ \ \ \ \ \ \ \ \ \ \ \ \ \ \ \  \ \  \ \ \ \ \ \ \ \ \ \ \ \ \ \ \ \ \ \ \ \ \ \ \ \ \ \ \ \ \ \ \ \ \  \ \  \ \ \ \ \ \ \ \ \ \ \ \ \ \ \ \ \ \ \ \ \ \ \ \ \ \ \  }\\
    \text{\textbf{One sentence:} A woman entered the kitchen and sliced a cucumber. \ \ \ \ \ \ \ \ \ \  \ \ \ \ \ \ \ \ \ \ \ \ \ \ \ \ \ \ \ \ \ \ \ \ \ \ \ \  \ \ \ \ \ \ \ \ \ \ \ \ \ \ \ \  \ \ \ \ \ \ \ \ \ \ \  \ \ \ \ \ \ \ \ \ }\\
    \end{array}$
    \caption{Sample annotation for TACoS-MultiLevel dataset.}
    \label{fig:TACOSM}
\end{figure}

From a background study, Senina et. al., \cite{rohrbach2014coherent} found that very less work has been done on generating multi-sentence linguistically coherent description. So, they have extended the annotation of the TACoS dataset by generating three levels of description for each video which includes: (1) a detailed paragraph of description with at max 15 sentences, (2) a 3-5 sentences short paragraph describing the videos event briefly and (3) a single sentence caption for the video. Similar to TACoS dataset \cite{tacos:regnerietal} this dataset also has 127 cooking videos with 26 cooking activities. Figure \ref{fig:TACOSM} shows the sample multi-sentence annotation of the TACoS-MultiLevel dataset.  

\subsubsection{YouCook II (2017)} For addressing the problem of Procedure Segmentation which is defined as the automatic segmentation of the video into independent procedure segments, Zhou et. al., \cite{zhou2017towards} \footnote{\url{http://youcook2.eecs.umich.edu/}} proposed a domain-specific cooking video dataset that included all the open domain challenges (such as different camera action and dynamic background). This dataset includes 2000 videos describing 89 different recipes. In this dataset, all the videos are annotated with temporal boundaries and contain 3-16 procedure segments per video. The length of the segments in the dataset varies between 1-264 seconds and has a vocabulary of 2600 words. The separation of the dataset is done randomly in the ratio of 67\%:23\%:10\% for training, validation, and testing. This dataset also includes some unseen recipes which are not included in the previous datasets. 

\subsection{Open domain datasets}
These datasets contain the videos from multiple domains such as movie, YouTube and vines etc. These datasets are more challenging than the domain-specific dataset due to various factors such as the involvement of different camera actions, collection of videos from different domains and most of the videos contain real-world challenges. Further, in this section, we will discuss different open domain datasets available for video description with their pros and cons. 

\subsubsection{Microsoft Video Description Dataset (MSVD, 2011)} 
\begin{figure}[!ht]
    \centering
$\begin{array}{cccccc}
    \includegraphics[scale=0.24]{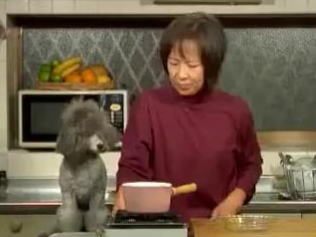}&    \includegraphics[scale=0.24]{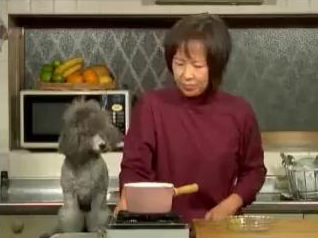}&   
    \includegraphics[scale=0.24]{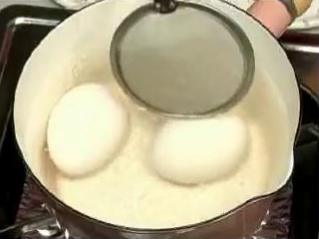}&   
    \includegraphics[scale=0.24]{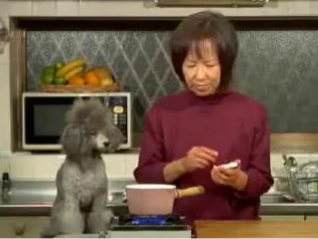}&    \includegraphics[scale=0.24]{msdv4.png}&    \includegraphics[scale=0.26]{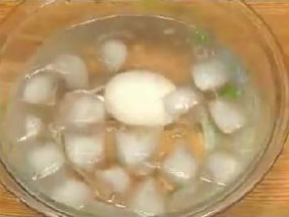}\\
   \multicolumn{3}{c}{\textbf{Ref1: }\text{the lady put two eggs into the water to boil. }} &
      \multicolumn{3}{c}{\textbf{Ref2: }\text{ a woman is boiling eggs in a pot of water. }} \\
   \multicolumn{3}{c}{\textbf{Ref3: }\text{a woman is boiling to eggs in a saucepan.\ \ \ \ \ }} &
      \multicolumn{3}{c}{\textbf{Ref4: }\text{ a woman turns two eggs in boiling water. }} \\
\end{array}$
    \caption{Sample annotation for MSVD dataset with selected reference caption out of 41 captions.}
    \label{fig:msdvds}
\end{figure}

On observing, most of the datasets available for video descriptions are domain-specific (especially cooking videos). Chen and Dolan \cite{chen2011collecting} released a standard MSVD\footnote{Available for download at \url{http://research.microsoft.com/en-us/downloads/38cf15fd-b8df-477e-a4e4-a4680caa75af/}} dataset of 1,970 YouTube videos with associated short natural language descriptions. Amazon’s Mechanical Turk is used for the collection of the dataset. Before releasing the dataset, manual filtration is performed to ensure that all the videos should meet the prescribed criteria and do not contain inappropriate and ambiguous content. To avoid bias lexical choice, the audio of all videos is muted. The domain of the annotation for the video is not limited to English. This dataset includes a multi-lingual natural description such as Chinese, English, German etc. The training, validation, and testing set of the dataset contains 1,200, 100 and 670 videos respectively. The dataset includes on average 41 single sentence descriptions for each video. Figure \ref{fig:msdvds} shows the sample annotation of the MSVD dataset with selected reference captions.

\subsubsection{MPII Movie Description Corpus (MPII-MD, 2015)}
MPII-MD dataset \cite{rohrbach2015dataset} consists of 94 Hollywood movies. For corresponding natural descriptions, transcribed audio descriptions are used. The audio description from Descriptive Video Service (DVS) is retrieved using MakeMkv\footnote{\url{ https://www.makemkv.com/}} and Subtitle edit\footnote{\url{http://www.nikse.dk/SubtitleEdit/}} and then the audio stream is transcribed using a crowd-sourced transcription service \cite{zhou2013crowdsourcing}. To overcome the misalignment between visual content and the time of the speech, manual alignment process for each sentence is employed. During this alignment process, the filtering of the descriptions is also carried out in which some descriptions are removed which describes the visual contents which are not present in the video, text extracted from the screen and a sentence describing movie introduction. All the 94 Hollywood movies are subdivided into 68,337 clips with corresponding 68,375 aligned descriptions with almost one sentence per clip.

\subsubsection{Montreal Video Annotation Dataset (M-VAD) 2015} This dataset contains the movie clips with associated natural language description narrating the visual actions and events in the clip. The dataset is based on Descriptive Video Service (DVS) which is a US-based descriptive narrations provider of movies and TV programs in DVDs and other media devices. M-VAD dataset \cite{torabi2015using} includes 48,986 video clips from 92 different movies with corresponding 55,904 sentences. The average length of clips in the dataset is 6.2 seconds and the length of the vocabulary is 17,609. The official split of the dataset is 38,949, 4,888, and 5,149 video clips for training, validation, and testing.

\subsubsection{Microsoft Research- Video to Text (MSR-VTT, 2016)} 
\begin{figure}[!ht]
    \centering
$\begin{array}{ccccc}
    \includegraphics[scale=0.20]{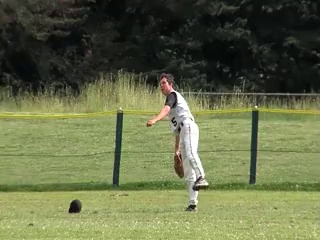}&    \includegraphics[scale=0.20]{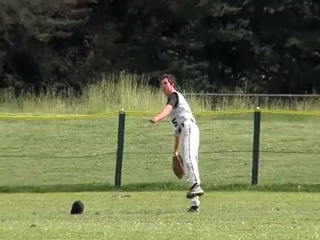}&    \includegraphics[scale=0.20]{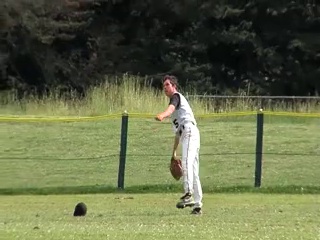}&    \includegraphics[scale=0.20]{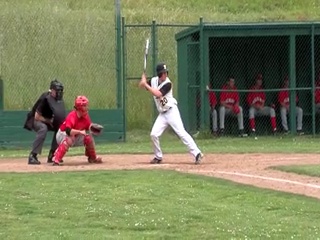}&    \includegraphics[scale=0.20]{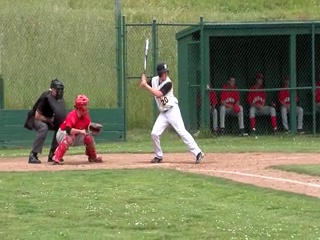}\\
\multicolumn{2}{c}{\textbf{Ref1: }\text{a baseball game is played.\ \ \ \ \ \ \ \ \  }}&\multicolumn{3}{c}{\textbf{Ref2: }\text{baseball player hits ball.\ \  \ \ \ \  \ \ \ \ \ \ \ \ \ \ \ \ \ \ \ \ \ \ \ \ \ \ \ \ \  \  }} \\
\multicolumn{2}{c}{\textbf{Ref3: }\text{a report about a baseball game.}}&\multicolumn{3}{c}{\textbf{Ref4: }\text{a batter hitting a ball and players on a field.}} \\
\end{array}$
    \caption{Sample annotation for MSR-VTT dataset with selected reference caption out of 20 captions.}
    \label{fig:msrvtts}
\end{figure}

As the research on video description starts attracting the interest of researchers, Xu et. al. \cite{xu2016msr}  released in 2016 a large dataset of 10K videos consisting of open domain videos. This dataset contains the videos of 20 different classes which are collected by analyzing 257 most used queries from a commercial video search engine. The separation of the dataset is done in such a way that the training, validation and test set contains 6,513, 497 and 2,990 videos respectively. In this dataset, there are 20 reference captions for each video which are annotated by AMT workers. This dataset is considered a standard dataset. It is one of the largest datasets in terms of the number of videos and associated reference captions. For the generation of the more fluent and descriptive natural description of the video, the audio of the videos is retained which helps various models in achieving state-of-the-art results. Figure \ref{fig:msrvtts} shows the sample annotation for MSR-VTT with selected reference captions.


\subsubsection{Video Titles in the Wild (VTW, 2016)} Consider the task of generating a catchy title for a video automatically in 2016 Zeng et. al. \cite{zeng2016generation} proposed a novel dataset. This dataset consists of 18,100 clips with only one corresponding sentence describing the video. The dataset is designed carefully in such a way that on average one word will appear in at most two sentences in the whole dataset. Along with a short single sentence describing the visual content of the video, this dataset also contains an augmented description that describes the visual contents which are not present in the video. Although the dataset is specifically proposed for title generation, it can also be used for other video-related tasks like video captioning, visual dialogue and question answering. 

\subsubsection{Charades (2016)} On observing performance of application of computer vision in assisting humans in managing their daily routine work, Sigurdsson et. al. \cite{sigurdsson2016hollywood} proposed a dataset in 2016. With the help of 267 AMT workers from three continents, this dataset includes 9,848 videos which include most of the daily indoor household activities. The approach of collecting this dataset is different from other datasets. In this dataset, initially a script is prepared which includes the description of the actions and object. Then, this script is given to workers and instructed to do the action with specified objects according to the script. For the selection of objects and actions, a fixed vocabulary is used. For the diversity in the location, 15 different indoor sites are selected. This dataset includes 27,847 annotations narrating 157 actions with 66,500 temporally localized intervals. The training and testing set contains 7,985 and 1,863 videos respectively. Further, for the video description, this raw dataset was refined by Wang et. al. \cite{wang2018video}.

\subsubsection{ActivityNet Captions (2017)} A video consists of more than one event and actions. With the research oriented motivation to address these situations on video description generation, Krishna et. al. proposed an ActivityNet Captions dataset \cite{krishna2017dense} in 2017. This dataset contains 20K videos which are the highest among all the datasets and include 100K dense reference descriptions. All the descriptions in the dataset are temporally localized with respect to events in the videos. Each description contains on average of 13.48 words. This dataset also gives multiple descriptions for each video which describe 94.6\% of the visual action and events in the video when combined. This dataset includes 10\% of temporal description overlapping for the events which occur simultaneously making it more challenging than other datasets.    

\subsubsection{ActivityNet Entities (2019)} 
The ActivityNet Entities\footnote{\url{Available at: https://github.com/facebookresearch/ActivityNet-Entities }}  dataset \cite{zhou2019grounded} contains the videos from training and validation set of ActivityNet captions \cite{krishna2017dense} since the ground truth for the test is not available publicly. The ActivityNet Entities dataset also includes the grounding and annotations of entities available in the frames of each video which makes this dataset diverse and more informative than other datasets. In this dataset, each frame is annotated with noun phrases from reference descriptions with bounding boxes. This dataset includes 14,281 annotated videos with around 52K video segments and 158K annotated bounding boxes. The training, validation, and test set contains 10K, 2.5K and 2.5K videos respectively.

\subsubsection{Video Story (2018)} For the generation of natural language narration of long social media videos, Gella et. al. \cite{gella2018dataset} proposed a multi-sentence description dataset in 2018. Unlike ActivityNet Captions \cite{krishna2017dense} or cooking video datasets \cite{zhou2017towards} where all the human activities are pre-decided. This dataset contains the activities from multiple domains. This dataset includes 20K videos and 26,245 paragraphs of natural language description which includes 123K sentences with an average of 13.32 words per sentence. The number of videos in training, validation and test set is 17,908, 999 and 1,011 videos respectively and a blind test set is also released consisting of 1,039 videos. In this dataset, there is a corresponding descriptive paragraph for each video in the training set and there are three paragraphs for output analysis for the validation and test set videos. The ground truth of the blind test is not released and it is only available on the server.

\begin{figure}[!ht]
    \centering
$\begin{array}{ccccc}
    \includegraphics[scale=0.10]{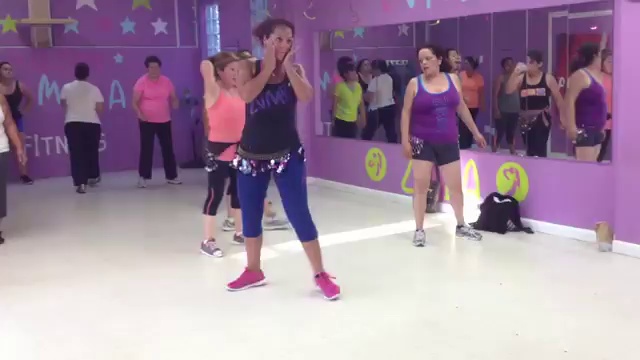}&    
    \includegraphics[scale=0.10]{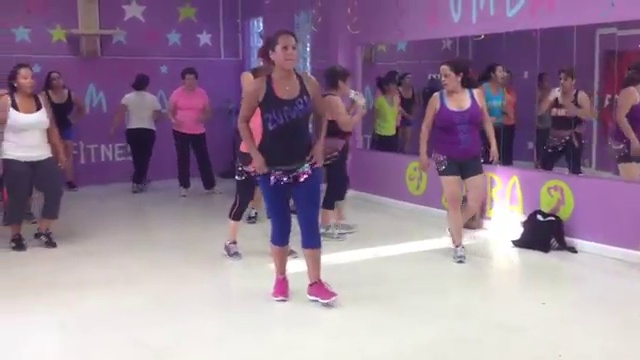}&    
    \includegraphics[scale=0.10]{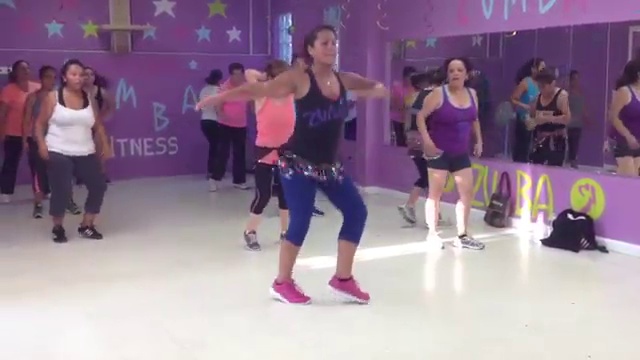}&    
    \includegraphics[scale=0.10]{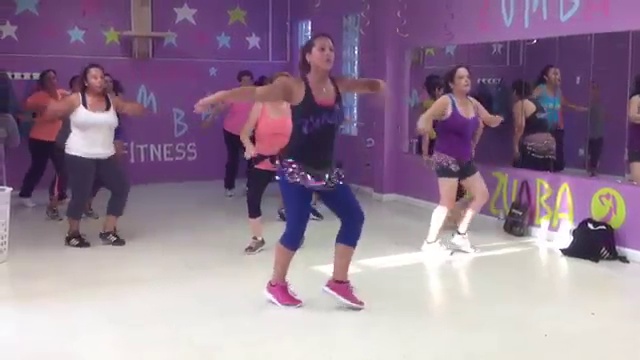}&
    \includegraphics[scale=0.10]{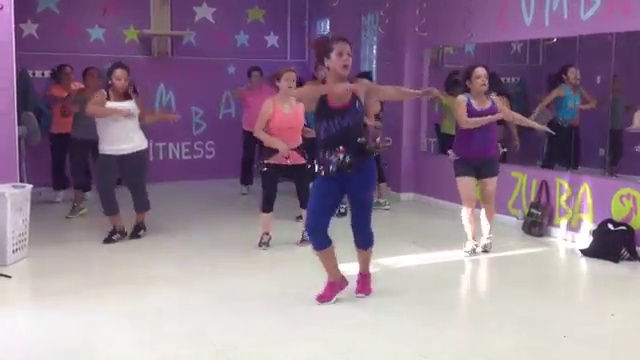}\\
\multicolumn{5}{c}{\textbf{Ref1: }\text{A group of woman are doing exercises to music in a workout studio. }} \\
\multicolumn{5}{c}{\textbf{Ref2: }\text{Some woman in a dance class dance to some spanish music.\ \ \ \ \ \ \ \ \ \ \ \  \ \ \ }} \\
\end{array}$
    \caption{Sample annotation from VATEX dataset with few English reference captions out of 20 English-Chinese captions.}
    \label{fig:vtx}
\end{figure}

\subsubsection{Video to Text (VATEX, 2019)} The video to text dataset (VATEX) is presented in 2019 for minimizing the gap between the language and vision. On observing, most of the datasets available for video description are in English and motivated the research on video guided MT. Wang et. al. \cite{wang2019vatex} introduces a multilingual dataset that includes 41,250 unique videos and 825,000 descriptions in both English and Chinese. Each video has corresponding 10 English and Chinese captions. Based on this dataset, two open challenge video to text (VATEX) and video guided machine translation (VMT) are organized in two different workshops in 2020. This dataset includes the largest number of video-sentence pairs with 10 English and Chinese natural descriptions for each video in the dataset. It includes 600 diverse activities in the whole dataset. The training, validation and test set of this dataset include 25991, 3000 and 6000 videos respectively. Apart from this separation, one secret set of 6,287 videos also introduced but the ground truth for this set is only available at the server\footnote{\url{https://competitions.codalab.org/competitions/24360}} for evaluation. Figure \ref{fig:vtx} shows the sample annotation for VATEX dataset and the Table \ref{tab:datasetd} gives the statistic of all the datasets.

\begin{table}[!ht]
    \centering
        \caption{Detail statistics of all the benchmarks dataset for video description.}
    \label{tab:datasetd}
    \begin{tabular}{|l|c|c|c|c|c|c|c|}\hline
\multirow{2}{*}{\textbf{ Dataset}} &\multirow{2}{*}{\textbf{Year}}& \multirow{2}{*}{\textbf{Domain}} & \multirow{2}{*}{\textbf{\#Classes}} & \multirow{2}{*}{\textbf{\#Clips}} & \multirow{2}{*}{\textbf{\#Sentence}} &\multirow{2}{*}{\textbf{Vocab}}&{\textbf{Multiple}} \\
       &&&&&&&\textbf{ references} \\\hline
       MSVD \cite{chen2011collecting}&2011&Open&218&1,970&70,000&13,010&\checkmark \\\hline       
       MP-II Cooking \cite{rohrbach2012database}&2012 &Cooking& 65&44&5609&-&$\times$\\\hline
       YouCook \cite{das2013thousand}&2013&Cooking&6&88&2,699&2,711&$\times$ \\\hline
       TACoS \cite{tacos:regnerietal}&2013&Cooking&26&7,206&18,277&28,292&$\times$ \\\hline
       TACoS-MultiLevel \cite{rohrbach2014coherent}&2014&Cooking&1&14,105&52,593&-&\checkmark\\\hline
       MPII-MD \cite{rohrbach2015dataset} &2015&Movie&-&68,375&68,375&24,549&$\times$ \\\hline
       M-VAD \cite{torabi2015using} &2015&Movie&-&48.986&55,904&17,609&$\times$ \\\hline
       MSR-VTT \cite{xu2016msr}&2016&Open&20&10,000&200,000&29,316&\checkmark \\\hline
       VTW \cite{zeng2016generation}&2016&Open&-&18,100&44,613&-&$\times$ \\\hline
       Charades  \cite{sigurdsson2016hollywood}&2016&Human&157&9,848&27,847&-&\checkmark \\\hline
       YouCook II \cite{zhou2017towards}&2017&Cooking&89&15.4K&15.4K&2,600&$\times$ \\\hline
       ActivityNet Captions \cite{krishna2017dense}&2017&Open&-&20,000&100,000&-&$\times$ \\\hline
       Video Story \cite{gella2018dataset}&2018&Social Media&-&123K&123K&-&\checkmark \\\hline
       ActivityNet Entities \cite{zhou2019grounded}&2019&Social Media&-&52,000&-&-&$\times$ \\\hline
       VATEX \cite{wang2019vatex}&2019&Open&-&41.3K&826K&-&\checkmark \\\hline

    \end{tabular}
\end{table}

\subsection{Open Video Description Challenges}
The main factor behind the success of the natural language description generation task is the active participation of various research groups, companies and conferences by introducing various challenges and competition. In this section, a brief discussion on various video description competitions is done.

\subsubsection{Large Scale Movie Description Challenge (LSMDC, 2015)}
The Large Scale Movie Description Challenge is organized as a event in conjunction with ICCV 2015, ECCV 2016, and ICCV 2017 and most recently with ICCV 2019. The challenge releases a blind test set along with publicly available training, validation and test set. The major tasks of this challenge are Annotation, Fill-in-the-Blank, and MovieQA. The dataset for LSMDC\footnote{The results of LSMDC challenge are available at \url{https://competitions.codalab.org/competitions/6121}} challenge is a combination of two previous standard datasets M-VAD \cite{torabi2015using} and MPII-MD \cite{zhou2013crowdsourcing}. This dataset includes 118,081 clips from 202 movies with one associated natural language description. The dataset is splitted into 91908, 6542 and 10053 videos for training, validation and testing. The blind test set includes 9578 video clips. For the evaluation of state-of-the-art models, the evaluation metrics used for image captioning are used.

\subsubsection{Microsoft Research- Video to Text (MSR-VTT, 2016)}
Further, in 2016 Microsoft introduced a competition named Microsoft Research - Video to Text (MSR-VTT) for reducing the gap between vision and languages. For this competition, the MSR-VTT dataset is used \cite{xu2016msr}. The main task is to generate a brief natural language description for a video. In the MSR-VTT dataset, there are 20 captions per video. The training, validation and test set includes 6,513, 497 and 2,990 videos respectively.

\subsubsection{Text Retrieval Conference (TREC, 2016)}
Text Retrieval Conference is a series of workshops started in early 2001 focusing on visual content analysis of videos.  Every year, this conference introduced new tasks. In 2016 this conference introduced a challenge named Video to Text (VTT) whose primary goal is to describe the video using natural language. Further, in 2017 again, this VTT task is organized in a dataset of 50K Twitter vines video. Based on the number of available descriptions (2-5), the dataset is divided into groups (G2, G3, G4, G5). 

\subsubsection{ActivityNet Challenge (2017)}
In 2017, a workshop at the CVPR conference is introduced a dense video captioning event as a task in ActivityNet Large Scale Activity Recognition Challenge \cite{ghanem2017activitynet}. The main goal of this task was to address most of the visual actions and events in the video in a form of natural language description. For this task, a dataset \cite{krishna2017dense} was introduced in which each video is annotated with a corresponding natural language description. All these descriptions are temporally aligned according to the events in the video. For the evaluation of the model, an online server\footnote{The results are available at \url{http://activity-net.org/challenges/2017/evaluation.html}} is evaluated using automatic metrics like BLEU, METEOR and CIDEr.

\subsubsection{Video to Text (VATEX, 2020)} In conjunction with CVPR 2020, a workshop on language and vision introduced a task for video understanding in two languages viz., English and Chinese. For this task, the VATEX dataset \cite{wang2019vatex} introduced in 2019 is used. This challenge\footnote{Available at \url{https://eric-xw.github.io/vatex-website/captioning_2020.html}} also introduced a blind test set for the evaluation of benchmark models along with training, validation and test set. The blind test set includes 6,278 videos with 10 English and Chinese captions. The training, validation and test set includes 25991, 3000 and 6000 videos respectively. For the evaluation of the generated output, MS-COCO Toolkit\footnote{\url{https://github.com/tylin/coco-caption}} provided by Microsoft for the evaluation of captions is used. 

\section{Video Description Evaluation Metrics and Benchmark Results}
In this section, a brief discussion is carried out on the evaluation metrics used for validating the performance of video description frameworks. In the second part, a comparative report on the recent benchmark results on each dataset is presented.

\subsection{Evaluation Metrics}
The natural language description of the video is evaluated either by employing a human evaluation approach or automatic evaluation metric. Both the human evaluation approach or automatic evaluation metrics have their pros and cons. Although there are no specific automatic evaluation metrics designed for the evaluation of the generated natural language description of the video, some of the most used automatic evaluation metrics are discussed in Section \ref{aem}. Further due to the unsatisfactory performance of the automatic evaluation metrics, human evaluation is performed as discussed in Section \ref{hme}.

\subsubsection{Automatic Evaluation}\label{aem}
The automatic evaluation of the output which is mainly in the form of natural language is challenging and when it comes to evaluating the generated natural language description of the video it gets more challenging due to the presence of multiple right answers. Due to the dynamic nature of the video and the presence of multiple actions or events, a video can be described by a wide variety of sentences. The structure of these sentences may vary syntactically and semantically. This can be seen in Section \ref{datas} where for each video there is multiple ground truth whereas for some video the same content is described multiple ways either by using synonyms or by employing different syntactic rules. 

Most of the automatic evaluation metrics which are used for the evaluation of the generated description are borrowed from the field of machine translation, text summarization and image captioning. The machine translation and text summarization evaluation metrics that are used for video description evaluation are Bilingual Evaluation Understudy (BLEU) \cite{papineni2002bleu}, Metric for Evaluation of Translation with Explicit Ordering (METEOR) \cite{banerjee2005meteor}, and Recall Oriented Understudy of Gisting Evaluation (ROUGE) \cite{lin2004rouge} respectively. Consensus-based Image Description Evaluation (CIDEr) \cite{vedantam2015cider}, and Semantic Propositional Image Captioning Evaluation (SPICE) \cite{anderson2016spice}. These are the metrics which are proposed for the evaluation of image captioning output but they are also used for evaluation of video description. Further, all the automatic evaluation metrics are briefly discussed with their merits and demerits.

\paragraph{\textbf{Bilingual Evaluation Understudy (BLEU)}}
BLEU is basically provided for the evaluation of machine translation output in 2002 by Papineni et. al. \cite{papineni2002bleu}. This evaluation metric basically looks for overlap of words between the predicted output and ground truth in terms of unigram, bigram, trigram and n-gram matching. In this metrics, as the score of higher n-gram matching increases, then it is assumed the output is closer to ground truth. Getting a BELU score of $1$ means that there is an exact matching of words between prediction and ground truth along with the order of occurrence. Equation \ref{eq:bp} is used for calculating the BLEU score.
  \begin{equation}\label{eq:bp}
  \begin{gathered}
      BP = {min(1, \exp(1-\frac{|reference|}{|output|}))} \quad\text{and} \quad
     n-gram~score = \sum_{n=1}^{N}w_{n}log p_n \quad   \text{Then,}\quad
        BLEU = BP+ n-gram~score
  \end{gathered}
  \end{equation} 
 


           

The Equation \ref{eq:bp} consists of three parts BP, n-gram score and BLEU. BP corresponds to Bravity Penalty which is basically used to penalize the prediction whose length is shorter. n-gram score is the geometric average of n-gram precision up to length $N$ ($p_n$) weighted by positive weight $w_n$ (sum of all weights is equal to $1$). In most of the baseline models, the value of $N$ is set to $4$ and $w_n$ to $\frac{1}{N}$. Although the BLEU score is very intuitive and boosted the natural language output evaluation process still has few shortcomings as it does not consider the synonyms of words. The BLEU score will be low if the number of references is $1$. So, the higher the number of reference captions higher the score we get. It is precision-oriented evaluation metrics that is why BP is used. At the end, this metric is basically designed to evaluate the corpus level text. Therefore, use for the evaluation of single sentence is not fair.

\paragraph{\textbf{Metric for Evaluation of Translation with Explicit Ordering (METEOR)}} METEOR \cite{banerjee2005meteor} is another evaluation metric from the field of machine translation which is basically proposed to address the issues of the BLEU evaluation metric. METEOR addresses the main issues of BLEU which does not consider the morphological variants of the words. In METEOR, the score is computed using unigram-precision, unigram-recall and a measure of fragmentation which is basically designed to show how well ordered the matched words are. METEOR supports matching at different levels such as stem level matching, synonyms level matching and exact word matching which makes it more suitable metrics for evaluating the natural description of the video. In the METEOR, firstly, an alignment is done between the pair of reference description and generated description. In this, every unigram in one description is mapped to the zero or one unigram in the corresponding description in the pair. This alignment process consists of a series of stages and on the completion of these stages, METEOR score is calculated.

 \begin{equation}\label{eq:mp}
     \begin{gathered}
       P = \frac{n_{rg}}{n_{g}} \quad R =\frac{n_{rg}}{n_{r}}\quad \text{and} \quad F-score = \frac{10PR}{R+9P}
\end{gathered}
 \end{equation}

Firstly, unigram precision (P) is the ratio of the number of unigrams that are present in both references and generated descriptions to the total number of unigrams present in the predicted description which is given in Equation \ref{eq:mp}. Then, a unigram recall is calculated as the ratio of the number of unigrams that are present in both reference description and generated description to the number of unigrams present in reference description as shown in Equation \ref{eq:mp}. Finally, F-score is calculated using a harmonic mean of recall and precision where more weightage is given to recall as shown in Equation \ref{eq:mp}. The recall, precision and F-score are totally based on unigram matching for considering longer matches. To calculate the penalty(p), all the unigrams are grouped into the minimum number of chunks. This is based on the intuition that the longer the n-gram match fewer chunks will be produced (for example if there is an exact matching between the ground truth and generated output then there will be only one chunk). All the unigrams in the chunks are adjacent in the reference description as well as in the generated description. The penalty and final METEOR score is calculated using Equation \ref{eq:pen}. 
\begin{equation}\label{eq:pen}
\begin{gathered}
   p = 0.5 \times \left( \frac{\#chunk}{\#unigrams\_matched}\right)^3 \quad
   M-score =  (1-p)\times F-score
\end{gathered}
\end{equation}



\paragraph{\textbf{Recall Oriented Understudy of Gisting Evaluation (ROUGE)}} ROUGE \cite{lin2004rouge} is another automatic evaluation metric which is basically proposed for the evaluation of text summarization output in 2004. Unlike BLEU which is a precision-oriented automatic evaluation metric, ROUGE is a recall-based evaluation metric that measures the number of n-gram overlap between the generated translation and the ground truth. There are five variants of the ROUGE that are proposed named as ROUGE-N (N-gram Co-Occurrence Statistics), ROUGE-L (Longest Common Subsequence (LCS)), ROUGE-W (Weighted Longest Common Subsequence), ROUGE-S (Skip-Bigram Co-Occurrence  Statistics) and ROUGE-SU (Extension of ROUGE-S). For the video description evaluation, ROUGE-L is used. ROUGE-L lookS for the longest common subsequence between generated natural language description and reference description and computes the recall and precision score. The intuition of ROUGE-L is that getting a longer LCS between the reference and generated sentence means the possibility of high similarity between the two summaries. The recall, precision and F-score as shown in Equation \ref{rcp} are computed to find how much similarity is between sentence $X$ of length $m$ and sentence $Y$ of length $n$.

\begin{equation}\label{rcp}
    \begin{gathered}
      R_{LCS}= \frac{LCS(X,Y)}{m}\quad P_{LCS}= \frac{LCS(X,Y)}{n}\quad\text{and}\quad
      F-score_{LCS}=\frac{(1+\beta^2)R_{LCS}P_{LCS}}{R_{LCS}+\beta^2P_{LCS}} \quad \text{where} \quad \beta = \frac{P_{LCS}}{R_{LCS}}
\end{gathered}
\end{equation}

Where $LCS(X, Y)$ is the longest common subsequence length between $X$ and $Y$ and the final value of $F-score_{LSC}$ will be the ROUGE-L score. The few advantages of ROGUE are that it does not look for consecutive matching but it employs in sequence matching within a sentence. In this, predefined length of n-gram is also not necessary since it automatically includes longest in-sequence common n-gram.

\paragraph{\textbf{Consensus-based Image Description Evaluation (CIDEr)}}
CIDEr \cite{vedantam2015cider} is the first evaluation metric that is specifically proposed for the evaluation of the image captioning output. It measures how well the generated description matches the consensus of the reference description. In this metric, firstly, stemming is performed that is mapping of all the words to their root form and then all the sentences are represented with a set of n-grams. The CIDEr score is based on the intuition that the consensus between the reference and generated description is encoded as the frequency of n-grams co-exist in both the descriptions. It also takes into consideration that the n-grams which are not present in the reference description should not occur in the generated description. All the n-grams are given a weightage based on the assumption that those n-grams whose occurrence is very common in the reference description of all the images are given low weightage since they carry less information about the visual content of the image. For assigning weights to all n-grams, Term Frequency Inverse Document Frequency (TF-IDF) \cite{robertson2004understanding} is used. The TF-IDF weighting $g_k(s_{ij})$ is calculated using Equation \ref{eq:tf} where $s_{ij}$ is a reference description where $h_{k}(s_{ij})$ denotes frequency of n-gram occurrence, $\Omega$ is n-grams vocabulary and $I$ is the set of images.
\begin{equation}\label{eq:tf} 
    g_k(s_{ij}) = \frac{h_{k}s_{ij}}{\sum_{w_l\in \Omega} h_l(s_{ij})}log\left( \frac{|I|}{\sum_{I_p\in I}min(1, \sum_{q}h_{k}s_{pq})}\right)
\end{equation}

\begin{equation}\label{eq:cdr}
\begin{gathered}
    CIDEr (c_i,S_i) = \sum_{n=1}^{N}w_{n}CIDEr_{n}(c_{i},S_{i})\quad \text{where} \quad CIDEr_{n}(c_{i},S_{i}) = \frac{1}{m}\sum_{j}\frac{g^{n}(c_{i})\cdot g^{n}(s(ij))}{||g^{n}(s_{ij}||\cdot {||g^{n}(s_{ij})||})} 
\end{gathered}
\end{equation}

The overall CIDEr score is calculate using Equation \ref{eq:cdr}, where $c_i$ is the generated description, $w_n$ is uniform weight given by $\frac{1}{N}$ where $N=4$, $g^{n}(c_i)$ is vector representing n-grams of all generated descriptions and ${||g^{n}(c_i)||}$ is magnitude of vector $g^{n}(c_i)$. 

\paragraph{\textbf{Semantic Propositional Image Captioning Evaluation (SPICE)}} SPICE is another automatic evaluation metric which is proposed for the evaluation of generated descriptions of the visual entities in an image after observing the incapabilities of the previous evaluation metrics which mainly focuses on n-gram overlapping. In the SPICE, both the reference and generated descriptions are transformed to a graphical semantic representation (known as scene graph) and then the similarity is measured. In this, all attributes, objects and their relationships are encoded by a dependency parse tree by a semantic scene graph. Let $c$ and $r$ be the generated and reference description respectively, then the scene graph $G(c)$ for generated description $c$ include object classes $O(c)$, relation types $R(c)$ and attribute types $A(c)$ (similarly for reference description $r$)where $G(c) = \langle x O(c),R(c),A(c)\rangle$. The final SPICE score (shown in Equation \ref{eq:rc}) is obtained by evaluating the F-score of similarity between the semantic scene graphs of generated and reference descriptions. The details of symbols used in Equation \ref{eq:rc} are given as follows. $T$ is a function that returns logical tuples of scene graphs, $\otimes$ is binary matching operator, $P$, $R$ and $F-score$ are precision, recall and F-score respectively.

\begin{equation}\label{eq:rc}
\begin{gathered}
   P(c,S)  = \frac{|T(G(c))\otimes T(G(S))|}{|T(G(c))|} \quad R(c,S) =  \frac{|T(G(c))\otimes T(G(S))|}{|T(G(S))|} \quad  SPICE(c,S) = F-score(c,S) = \frac{2\cdot P(c,S)\cdot R(c,S)}{P(c,S)+ R(c,S)} 
\end{gathered}
\end{equation}


\subsubsection{Human Evaluation}\label{hme}
The automatic evaluation metrics for the evaluation of natural language descriptions of visual content, action, events in the video are either borrowed or motivated from the field of natural language processing. Due to this, these automatic evaluation metrics lack in various fields such as dependency on n-grams overlapping reduces the score for the generated descriptions which are correct according to the visual content of the video but not according to reference description and theses evaluation metrics perform poorly when the number of reference captions for evaluation are not more than one. In the human evaluation process, a group of people who are experts in the field of video description analyze the generated descriptions and then report about the performance of the model in terms of coherency, relevancy, and diversity. Since the video has a dynamic content as compared to the image the evaluation process gets more challenging because there are many cases where the model has addressed some action or object which annotators have not addressed in the reference caption or they have addressed them in a different way. These types of situation motivate researchers to report human evaluation score along with the score of automatic evaluation metrics. The human evaluation also helps to analyze the generated description syntactically and semantically.  

\subsection{Shortcomings and  Reliability of Evaluation Metrics}
Whether it is a human evaluation or an automatic evaluation metric, both have their advantages or disadvantages. Till now, there is no specific evaluation metric that is specifically designed or proposed to evaluate the generated natural description for the video \cite{aafaq2019video}. The automatic evaluation metrics are computationally time-efficient and required less human effort. All the automatic evaluation metrics used to evaluate natural language description are the extended version of the metrics used for the evaluation of the output of machine translation and image captioning. These metrics have one common thing i.e., they all look for n-gram overlapping. Based on 'what is described' and `how it is described' a video can have multiple descriptions. So, for the evaluation of the model, all the possible reference descriptions are required. Several studies \cite{yu2016video,venugopalan2015sequence} have proved that METEOR, CIDEr and SPICE shows better performance in the evaluation of the natural language description in terms of human judgment. The human evaluation approach has also shown promising results although the human evaluation process gets time-consuming but still it gives a better analysis of generated description in terms of coherency, relevancy, and diversity. In the closure, as both approaches have merits and demerits, both human evaluation and automatic evaluation metrics should be employed together for the evaluation of generated descriptions until an efficient and effective evaluation metric is proposed specifically for the evaluation of video description.

\subsection{Results of State-of-the-art Approaches}\label{sec:bncrs}
In this section, the benchmark results of state-of-the-art techniques on all datasets are discussed. All the results are grouped according to the dataset for more clarity and comprehensibility. If there are multiple variants of the same model, then the results of the best variant are reported. For in-depth details of other variants of the model, the original paper should be consulted. For the comparative analysis of the models, we selected BLEU-4 from the available n-grams and METEOR, CIDEr, and ROUGE from the available evaluation metrics.

The results for the MSVD dataset are shown in Table \ref{tab:Rmsdv}. From Table \ref{tab:Rmsdv}, it is observed that till now there is no approach that is excelling in all the automatic evaluation metrics. The approach proposed in \cite{daskalakis2018learning} for learning spatiotemporal features for the video description achieved the highest BLEU-4 score of $61.30$ while \cite{zhang2020object} reported the highest METEOR, CIDEr, and ROUGE score of $36.10$, $95.20$, and $79.90$ respectively. Table \ref{tab:Rmsdv} also reports the results on the MSR-VTT dataset which is a challenging dataset for short video description. Unlike MSVD, the spatiotemporal features learning-based approach proposed in \cite{xiao2020video} achieved the highest benchmark results on the MSR-VTT dataset and outperforms other approaches with a respectable margin in BLEU-4 and ROUGE evaluation metrics. The reinforcement learning-based approach proposed in \cite{pasunuru2017reinforced} reports a CIDEr score of $51.70$ which is the highest till now among all the state-of-the-art approaches.  

\begin{table}[!ht]
    \caption{ Results of video description approaches on MSVD and MSR-VTT datasets}
    \label{tab:Rmsdv}
    \centering
    \begin{tabular}{|c|c|c|c|c|c|c|c|c|c|}\hline
    \multirow{2}{*}{\textbf{Methods}} & \multirow{2}{*}{\textbf{Year}}&\multicolumn{4}{c|}{\textbf{MSVD  Results}}&\multicolumn{4}{c|}{\textbf{MSR-VTT Results}}\\\cline{3-10}
&&\textbf{BLEU-4}&\textbf{METEOR}&\textbf{CIDEr}&\textbf{ROUGE}&\textbf{BLEU-4}&\textbf{METEOR}&\textbf{CIDEr}&\textbf{ROUGE}\\\hline
\cite{zhang2020object}&2020&54.30&\textbf{36.40}&\textbf{95.20}&\textbf{73.90}&43.60&\textbf{28.80}&{50.90}&{62.10}\\\hline
\cite{xiao2020video}&2020&54.10&{36.10}&86.10&{72.40}&\textbf{44.60}&{28.70}&48.60&\textbf{62.20}\\\hline
\cite{xu2020video}&2020&48.80&33.40&82.00&69.70&37.90&28.40&40.60&59.30\\\hline
\cite{wang2020sequence}&2020&56.10&34.90&88.20&-&-&-&-&-\\\hline
\cite{sah2020understanding}&2020&43.00&33.20&71.10&68.70&40.50&28.20&45.30&60.40\\\hline
\cite{wei2020exploiting}&2020&46.80&34.40&85.70&-&38.50&26.90&43.70&-\\\hline
\cite{liu2020sibnet}&2020&54.20&34.80&88.20&71.70&40.90&27.50&47.50&60.20\\\hline
\cite{nabati2020multi}&2020&53.30&33.80&76.40&-&39.50&27.50&42.80&-\\\hline
\cite{nabati2020video}&2020&42.90&32.00&62.20&68.30&36.60&27.00&40.50&58.70\\\hline
\cite{XiaoHuanhou}&2019&51.30&35.10&82.90&-&41.20&27.70&43.90&-\\\hline
\cite{aafaq2019spatio}&2019&47.90&35.00&78.10&71.50&38.30&28.40&48.10&60.70\\\hline
\cite{xu2019semantic}&2019&51.20&35.40&74.90&-&40.80&28.70&46.80&61.50\\\hline
\cite{gao2019fused}&2019&52.70&34.50&-&-&40.80&27.40&-&-\\\hline
\cite{li2019end}&2019&50.30&34.10&87.50&-&40.40&27.00&48.30&-\\\hline
\cite{9022152}&2019&50.00&34.30&86.60&70.20&40.50&27.70&47.60&60.60\\\hline
\cite{chen2019boundary}&2019&44.10&32.12&70.10&-&38.70&26.70&41.00&-\\\hline
\cite{daskalakis2018learning}&2018&\textbf{61.30}
&33.80&63.20&-&&&&\\\hline
\cite{li2018multimodal}&2018&48.00&31.60&68.80&-&37.50&26.40&-&-\\\hline
\cite{chen2018less}&2018&46.10&33.10&76.00&69.20&38.90&27.20&42.10&59.50\\\hline
\cite{wang2018m3}&2018&52.80&33.30&-&-&38.10&26.60&-&-\\\hline
\cite{wang2018reconstruction}&2018&52.30&34.10&80.30&69.80&39.10&26.60&42.70&59.30\\\hline
 \cite{wang2018video}&2018&-&-&-&-&41.30&{28.70}&48.00&-\\\hline
\cite{wu2018interpretable}&2018&51.70&34.00&74.90&-&-&-&-&-\\\hline
\cite{lee2018multimodal}&2018&50.00&-&{94.30}&-&41.80&-&-&-\\\hline
\cite{liu2017hierarchical}&2017&44.30&32.10&68.40&68.90&-&-&-&-\\\hline
\cite{nian2017learning}&2017&40.10&29.90&51.10&-&-&-&-&-\\\hline
\cite{gao2017video}&2017&50.80&33.30&74.80&-&38.00&26.10&-&-\\\hline
\cite{xu2017learning}&2017&52.30&33.60&70.40&-&36.50&26.50&41.00&59.80\\\hline
\cite{tu2017video}&2017&51.10&32.70&67.50&-&37.40&26.60&41.50&-\\\hline
\cite{pasunuru2017reinforced}&2017&54.40&34.90&88.60&-&40.50&28.40&\textbf{51.70}&-\\\hline
    \end{tabular}
\end{table}

Table \ref{tab:Rmvad} reports the results of various approaches on the M-VAD dataset. The M-VAD is a challenging dataset which can be concluded based on the observation from Table \ref{tab:Rmvad}. Till now no approach is able to report a high benchmark result for the M-VAD dataset. The recent sequence in sequence approach proposed in \cite{wang2020sequence} is also not able to outperform other models on the M-VAD dataset although the results are comparable to other approaches. \cite{xu2019semantic} report highest METEOR score of $8.3$.  Further, Table \ref{tab:Rmvad} also presents the score of state-of-the-art approaches on the MPII-MD dataset. It is another challenging dataset since it is basically proposed for generating natural language description for movie clips. Results on MPII-MD datasets also needs to be improved.

\begin{table}[!ht]
    \caption{ Results of video description approaches on M-VAD and MPII-MD datasets}
    \label{tab:Rmvad}
    \centering
     \begin{tabular}{|c|c|c|c|c|c|c|c|c|c|}\hline
    \multirow{2}{*}{\textbf{Methods}} & \multirow{2}{*}{\textbf{Year}}&\multicolumn{4}{c|}{\textbf{M-VAD Results}}&\multicolumn{4}{c|}{\textbf{MPII-MD Results}}\\\cline{3-10}
&&\textbf{BLEU-4}&\textbf{METEOR}&\textbf{CIDEr}&\textbf{ROUGE}&\textbf{BLEU-4}&\textbf{METEOR}&\textbf{CIDEr}&\textbf{ROUGE}\\\hline
\cite{wang2020sequence}&2020&&{7.3}&-&-&-&-&-&-\\\hline
\cite{sah2020understanding}&2020&\textbf{1.0}&{6.9}&-&--&-&-&-&-\\\hline
\cite{xu2019semantic} &2019&-&\textbf{8.3}&-&--&-&-&-&-\\\hline
\cite{nian2017learning}&2017&&5.7&-&-&-&6.60&-&-\\\hline
\cite{Baraldiet}&2017&-&{7.3}&-&-&\textbf{0.80}&7.00&\textbf{10.8}&16.7\\\hline
\cite{pan2017video}&2017&-&7.2&-&-&-&\textbf{8.00}&-&-\\\hline
\cite{liu2017hierarchical}&2017&-&-&-&-&0.60&7.10&8.90&\textbf{17.0}\\\hline
\cite{Pan:CVPR16}&2016&-&6.7&-&-&-&7.30&-&-\\\hline
\cite{14}&2016&-&6.8&-&-&-&6.80&-&-\\\hline
\cite{PanPingbo}&2016&{0.7}&6.8&-&-&-&-&-&-\\\hline
\cite{rohrbach2015dataset}&2015&-&-&-&-&-&5.60&-&-\\\hline
\cite{rohrbach2015long}&2015&-&6.4&-&-&-&7.00&-&-\\\hline
\cite{venugopalan2015sequence}&2015&-&6.7&-&-&-&7.10&-&-\\\hline
\cite{yao2015describing}&2015&{0.7}&5.7&\textbf{6.1}&-&-&-&-&-\\\hline
    \end{tabular}
\end{table}


 Table \ref{tab:RCharades} reports a result on the Charades dataset proposed in 2016. It contains the indoor activities performed by humans. In \cite{sigurdsson2016hollywood}, Sigurdsson et. al. found that the sequence to sequence model proposed in \cite{venugopalan2015sequence} reports the best result as a baseline model. Further in 2018 \cite{wang2018video}, reinforcement learning-based approach outperforms other previous models. The results proposed in \cite{wang2018video} are still state-of-the-art which can be seen in Table \ref{tab:RCharades}.

\begin{table}[!ht]
    \caption{ Results of video description approaches on Charades dataset}
    \label{tab:RCharades}
    \centering
    \tabcolsep 14pt
    \begin{tabular}{|c|c|c|c|c|c|}\hline
    \multirow{2}{*}{\textbf{Methods}} & \multirow{2}{*}{\textbf{Year}}&\multicolumn{4}{c|}{\textbf{Results}}\\\cline{3-6}
&&\textbf{BLEU-4}&\textbf{METEOR}&\textbf{CIDEr}&\textbf{ROUGE}\\\hline
\cite{wei2020exploiting}&2020&12.7&17.2&21.6&-\\\hline \cite{wang2018video}&2018&\textbf{18.8}&\textbf{19.5}&\textbf{23.2}&\textbf{41.4}\\\hline
\cite{wu2018interpretable}&2018&13.5&17.8&20.8&-\\\hline
\cite{venugopalan2015sequence}&2016&6.0&13.0&17.0&31.0\\\hline
    \end{tabular}
\end{table}

Table \ref{tab:RActivityNet Captions} reports the results on ActivityNet Captions dataset which is proposed for long video descriptions. The results on the ActivityNet Captions dataset are still low which may be due to the challenges present in the dataset.

Further, Table \ref{tab:Rvatex} reports the results on the recent VATEX dataset. \cite{wang2019vatex} reports the baseline results on this dataset. Other approaches such as \cite{singh2020nits}, \cite{lin2020multi}, and \cite{zhu2019multi} participated in the VATEX video captioning challenge 2020\footnote{\url{https://eric-xw.github.io/vatex-website/captioning_2020.html}} and report their results. The multi-features and hybrid reward strategy approach proposed in \cite{zhu2019multi} was the winner of the video captioning competition and reports the highest result on the VATEX dataset. Table \ref{tab:Rvatex} also shows the performance of the state-of-the-art approaches in generating Chinese captions for the VATEX dataset.

\blfootnote{The results for VATEX dataset can be found at \url{https://competitions.codalab.org/competitions/24360}}
\begin{table}[!ht]
    \caption{ Results of video description approaches on ActivityNet Captions dataset}
    \label{tab:RActivityNet Captions}
    \centering
    \tabcolsep 14pt
    \begin{tabular}{|c|c|c|c|c|c|}\hline
    \multirow{2}{*}{\textbf{Methods}} & \multirow{2}{*}{\textbf{Year}}&\multicolumn{4}{c|}{\textbf{Results}}\\\cline{3-6}
&&\textbf{BLEU-4}&\textbf{METEOR}&\textbf{CIDEr}&\textbf{ROUGE}\\\hline
\cite{xu2019joint}&2019&1.63&8.58&19.88&\textbf{19.63}\\\hline
\cite{wang2018bidirectional}&2018&2.30&9.60&12.68&19.10\\\hline
\cite{li2018jointly}&2018&1.62&10.33&\textbf{25.24}&-\\\hline
\cite{zhou2018end}&2018&2.23&9.56&-&-\\\hline
\cite{krishna2017dense}&2017&\textbf{3.98}&9.50&24.6&-\\\hline
\cite{yao2017msr}&2017&-&\textbf{12.84}&-&-\\\hline
\cite{chen2018ruc+}&2017&-&9.61&-&-\\\hline
    \end{tabular}
\end{table}


\begin{table}[!ht]
    \caption{ Results of video description approaches on VATEX English and  Chinese dataset}
    \label{tab:Rvatex}
    \centering
    \begin{tabular}{|c|c|c|c|c|c|c|c|c|c|}\hline
    \multirow{2}{*}{\textbf{Methods}} & \multirow{2}{*}{\textbf{Year}}&\multicolumn{4}{c|}{\textbf{VATEX English Results}}&\multicolumn{4}{c|}{\textbf{VATEX  Chinese Results}}\\\cline{3-10}
&&\textbf{BLEU-4}&\textbf{METEOR}&\textbf{CIDEr}&\textbf{ROUGE}&\textbf{BLEU-4}&\textbf{METEOR}&\textbf{CIDEr}&\textbf{ROUGE}\\\hline
\cite{zhu2019multi}&2020&\textbf{40.7}&\textbf{25.8}&\textbf{81.4}&\textbf{53.7}&{32.6}&\textbf{32.1}&\textbf{59.5}&\textbf{56.6}\\\hline
\cite{singh2020nits}&2020&22.0&18.0&27.0&43.0&-&-&-&-\\\hline
\cite{lin2020multi}&2020&39.2& 25.0&76.0& 52.7&\textbf{33.0}& 30.3&50.4&49.7  \\\hline  
\cite{zhang2020object}&2020&32.1&22.2&49.7&48.9&-&-&-&-\\\hline
\cite{wang2019vatex}&2019&28.5&21.6&45.1&47.0&24.8&29.4&35.1&51.6 \\\hline 
    \end{tabular}
\end{table}


\section{Future Research Directions and Challenges}
The third phase in the field of video description gains huge success in generating more fluent and descriptive natural language description for the video as compared to other phases due to the adoption of deep learning based approaches. This section includes a brief discussion on open challenges in the field of video description, the futures research directions, and the application of video description.

\subsection{Future Research Direction}
\subsubsection{Video Guided Machine Translation (VMT)}
As the multimodal machine translation gained popularity due to its success in generating better translations and disambiguation of words with multiple meanings, researchers also have started exploring the field of video guided machine translation. Various dedicated workshops on language and vision have introduced challenges based on VMT like   Advances in Language and Vision Research (ALVR 2020) workshop in ACL2020 organized a challenge on video guided machine translation\footnote{\url{https://eric-xw.github.io/vatex-website/translation_2020.html}} in English and Chinese. For that, they have introduced a large scale dataset \cite{wang2019vatex}. The intuition of using video guided machine translation is to take as much as benefits of all the diverse content in the video for translation. 

\subsubsection{Video Sentiment Analysis}
As the generation of multimedia data is increasing, especially video, the need for a system that automatically detects the sentiment of the video whether it is spreading any negativity and hate among the society also gained popularity. This makes the prediction of the sentiment of the video or the activity in the video as important as the prediction of the sentiment of the text. Till now, very few works have been reported in this field but in the near future, we can expect some work on this field as the researchers have already started exploring the field of multimodal sentiment analysis with the help of images.

\subsubsection{Visual Dialogue/Question Answering}
Similar to the audio dialogue, the visual dialogue has also attracted the attention of researchers. As compared to audio dialogue, the visual dialogue suffers more challenges. But, it has a broader area of applications such as enabling robots to be capable of having effective visual question answering and online chatbots systems. The idea of visual dialogues is that for a given video, the model should understand the actions, events, motion of the video and remember everything to give an answer to the users when asked. An effective and efficient video description generation model can help the Visual Dialogue/Question Answering model in generating some logical inferences from the generated natural language descriptions and participate in giving responses to the series of questions related to the visual content of the video. Some of the intelligent systems where a description of the visual content can play a major role are guiding self-driving cars, generating a natural language description of the action performed by the peoples in a surveillance camera and natural language description of the visual content can also help AI chatbots in taking more precise decisions.

\subsubsection{Video Indexing and Retrieval (VIR)}
The generation of a short description of video can also help in the field of automatic video indexing and retrieval. As the number of short videos is increasing over the internet, the process of generating tags and the categorization of the videos based on the content is getting challenging. An intelligent video description system can help in understanding the actions performed in the video and the relationships among the objects in the video and generate a natural language description from which we can select the appropriate keywords for the video.


\subsection{Open Challenges}
Different phases of video description approaches are analyzed and discussed in this work. From the survey, we observed that the present state-of-the-art approaches report important findings and display good contributions in this field. Although these approaches have taken the quality of generated natural description for the video to new heights. Still, there are some open challenges that need to be addressed for the flourishing of this research area. 

\subsubsection{Frame Selection}
A video contains a large number of similar frames since it is a collection of a sequence of continuous similar frames. Sampling these frames into chunks can lead the segment containing multiple events of the video together in a single segment which can affect the quality of extracted features. Moreover, it can create chaos in the successful classification of the event since the feature is extracted from a segment that contains more than one event. The existing video description frameworks sample the frames into chunks of 20 frames or 30 frames and extract the visual features from these segments. The major issues with these approaches are how effectively the video is segmented and the criterion for the dimension of the chunks. The sampling of frames does not guarantee that all selected frames in a segment are informative as they are prone to motion, illumination and occlusion effect \cite{chen2018less}. Although \cite{chen2018less} proposed an approach by using an average of 5-6 frames per video, this approach is outperformed by the state-of-the-art approaches. So, the selection of appropriate number of frames and keyframes from the video for visual feature encoding is still an unsolved problem. 

\subsubsection{Feature Selection} For generating a textual description of the video using a deep learning approach, selection of appropriate features plays a major role. Currently, lots of models have been proposed using multiple feature modalities such as object features, semantic features, motion features and audio features. Most of the approaches are dependent on pre-trained 3D CNN \cite{karpathy2014large} architectures for extracting the visual features. These pre-trained models have limitations like they at once can proceed only limited numbers of frames (15-16 frames). \cite{xu2019semantic} shows that audio and semantic feature plays a major role in recording better automatic scores but the current state-of-the-art approach in the MSR-VTT dataset uses 3D CNN features with dynamic attention and step-by-step learning which contradict the statement of \cite{xu2019semantic}. So, it is still challenging to propose an effective and computational efficient visual features encoder \cite{aafaq2019video} for the task of the video description.

\subsubsection{Temporal Attention for Text Generation} In the image caption generation, visual attention has been proved very effective in generating a text relevant to the visual entity. Similarly, here in video captioning, temporal attention can play a major role in generating a relevant text by selecting a segment of frames. Along with temporal attention, semantic attention also needs to be considered because for words like `the' and 'is', semantic attention can play a major role than visual temporal attention. The redundancy due to similar frames in a video is still an unsolved problem as all the frames are considered for generating text from the video which increases the chances that selecting irrelevant features directly affects the efficiency of temporal attention.

\subsubsection{Boundaries in the Video} Although various boundary aware approaches \cite{xu2019semantic, Baraldiet, Shinet, chen2018less, PanPingbo} have been proposed. But, in these approaches, only abrupt boundary (transition) changes are addressed and gradual boundary changes are ignored. These approaches also do not report the effectiveness of the boundary detection module as the boundary detection process is also prone to many challenges like illumination, motion and the occlusion effect. On analyzing various shot boundary detection based approaches, we observed that detecting gradual transition along with abrupt transition may help in removing non-informative frames which in result can help in selecting informative frames.

\subsubsection{Evaluation Metric} The major challenge that slows down the pace of research in the field of video description is the availability of an effective evaluation metric. Presently, all the evaluation metrics used for evaluation are borrowed from machine translation, text summarization and image captioning. These evaluation metrics perform well only when the number of references available is more than one. So, designing evaluation metrics specifically for the video description is a major bottleneck in the progress of this field.

\section{Conclusion}
In this survey, we reviewed all the phases of video description approaches with more focus on recent deep learning-based approaches and the availability of datasets with their pros and cons. We discussed the role of research from industries and research institutes in the growth of video description tasks by organizing several open challenges in this field. Although deep learning-based approaches report state-of-the-art results in the task of video description. Still, lots of work needs to be done like an automatic evaluation metric score for generating long descriptions or addressing most of the events in the long videos of MPII-MD, ActivityNet captions and M-VAD datasets which can be observed in Section \ref{sec:bncrs}. All the above discussed open challenges and their poor results in generating a long narration for the video makes the task of video description still an active research area where lots of things are there to be done. The active participation of researchers from industries and research institutes can overcome most of the challenges in near future. 

\section*{ACKNOWLEDGMENT}
This work is supported by Scheme for Promotion of Academic and Research Collaboration (SPARC) Project Code: P995 of No: SPARC/2018-2019/119/SL (IN) under MHRD, Govt of India.

\bibliographystyle{ACM-Reference-Format}
\bibliography{bib}

\end{document}